\pdfoutput=1
\documentclass[11pt]{article}

\usepackage[preprint]{acl}

\usepackage{times}
\usepackage{latexsym}
\usepackage[T1]{fontenc}
\usepackage[utf8]{inputenc}
\usepackage{microtype}
\usepackage{inconsolata}
\usepackage{graphicx}

\usepackage[inline]{enumitem}
\usepackage{caption}
\usepackage{subcaption}
\usepackage{times}
\usepackage{latexsym}
\usepackage[fleqn]{amsmath}
\usepackage{amssymb}
\usepackage{float}
\usepackage{algorithm}
\usepackage{algpseudocode}
\usepackage{soul}
\usepackage{booktabs}
\usepackage{tabularx}
\usepackage{multirow}

\usepackage{arydshln} 

\usepackage{tikz}
\usetikzlibrary{patterns}

\usepackage{placeins}

\usepackage{tcolorbox} 

\tcbset{
  graybox/.style={
    colback=lightgray!30, 
    colframe=white, 
    boxrule=0pt, 
    arc=2mm, 
    left=3pt, 
    right=6pt, 
    top=3pt, 
    bottom=3pt, 
    boxsep=0mm, 
    before skip=3pt, 
    after skip=3pt, 
  }
}

\tcbset{
  bluebox/.style={
    colback=blue!20, 
    colframe=white, 
    boxrule=0pt, 
    arc=2mm, 
    left=3pt, 
    right=6pt, 
    top=3pt, 
    bottom=3pt, 
    boxsep=0mm, 
    before skip=3pt, 
    after skip=3pt, 
  }
}

\DeclareRobustCommand{\redcircle}{\tikz[baseline=-0.6ex]\fill[red] (0,0) circle (0.6ex);}
\DeclareRobustCommand{\bluecircle}{\tikz[baseline=-0.6ex]\fill[blue] (0,0) circle (0.6ex);}
\DeclareRobustCommand{\greenpatterncircle}{\tikz[baseline=-0.6ex]{\draw[pattern=north west lines, pattern color=green] (0,0) circle (0.8ex); \draw[black] (0,0) circle (0ex);}}

\newcommand{\da}[1]{$\downarrow#1 $}
\newcommand{\ua}[1]{$\uparrow#1 $}

\usepackage[textsize=footnotesize]{todonotes}

\title{Class Distillation with Mahalanobis Contrast: An Efficient Training Paradigm for Pragmatic Language Understanding Tasks}

\author{Chenlu Wang $\qquad$ Weimin Lyu $\qquad$ Ritwik Banerjee \smallskip\\
         Department of Computer Science \\
         Stony Brook University, New York, USA}

\begin{document}
\maketitle
\begin{abstract}
Detecting deviant language such as sexism, or nuanced language such as metaphors or sarcasm, is crucial for enhancing the safety, clarity, and interpretation of online social discourse. While existing classifiers deliver strong results on these tasks, they often come with significant computational cost and high data demands. In this work, we propose \textbf{Cla}ss \textbf{D}istillation (ClaD), a novel training paradigm that targets the core challenge: distilling a small, well-defined target class from a highly diverse and heterogeneous background. ClaD integrates two key innovations: (i) a loss function informed by the structural properties of class distributions, based on Mahalanobis distance, and (ii) an interpretable decision algorithm optimized for class separation. Across three benchmark detection tasks -- sexism, metaphor, and sarcasm -- ClaD outperforms competitive baselines, and even with smaller language models and orders of magnitude fewer parameters, achieves performance comparable to several large language models (LLMs). These results demonstrate ClaD as an efficient tool for pragmatic language understanding tasks that require gleaning a small target class from a larger heterogeneous background.
\end{abstract}
\section{Introduction}
\label{sec:intro}
The widespread adoption of social media and the polarized nature of online discourse have amplified the need for improved communication dynamics, fostering research aimed at promoting safety and mutual respect. A critical part of this effort involves detecting complex linguistic phenomena such as figurative speech -- like sarcasm and metaphor~\cite{riloff2013sarcasm, oraby2016creating, ghosh2020report, ge2023survey} -- as well as harmful language patterns like aggressive rhetoric or sexism~\cite{samghabadi2020aggression, samory2021sexism}. These tasks present significant urgency and challenges due to the nuances of figurative speech and the variability of deviant language.

\begin{figure}[!t]
\centering
\frame{
\includegraphics[trim={1mm 4mm 4mm 15mm}, clip, width=.9\linewidth]{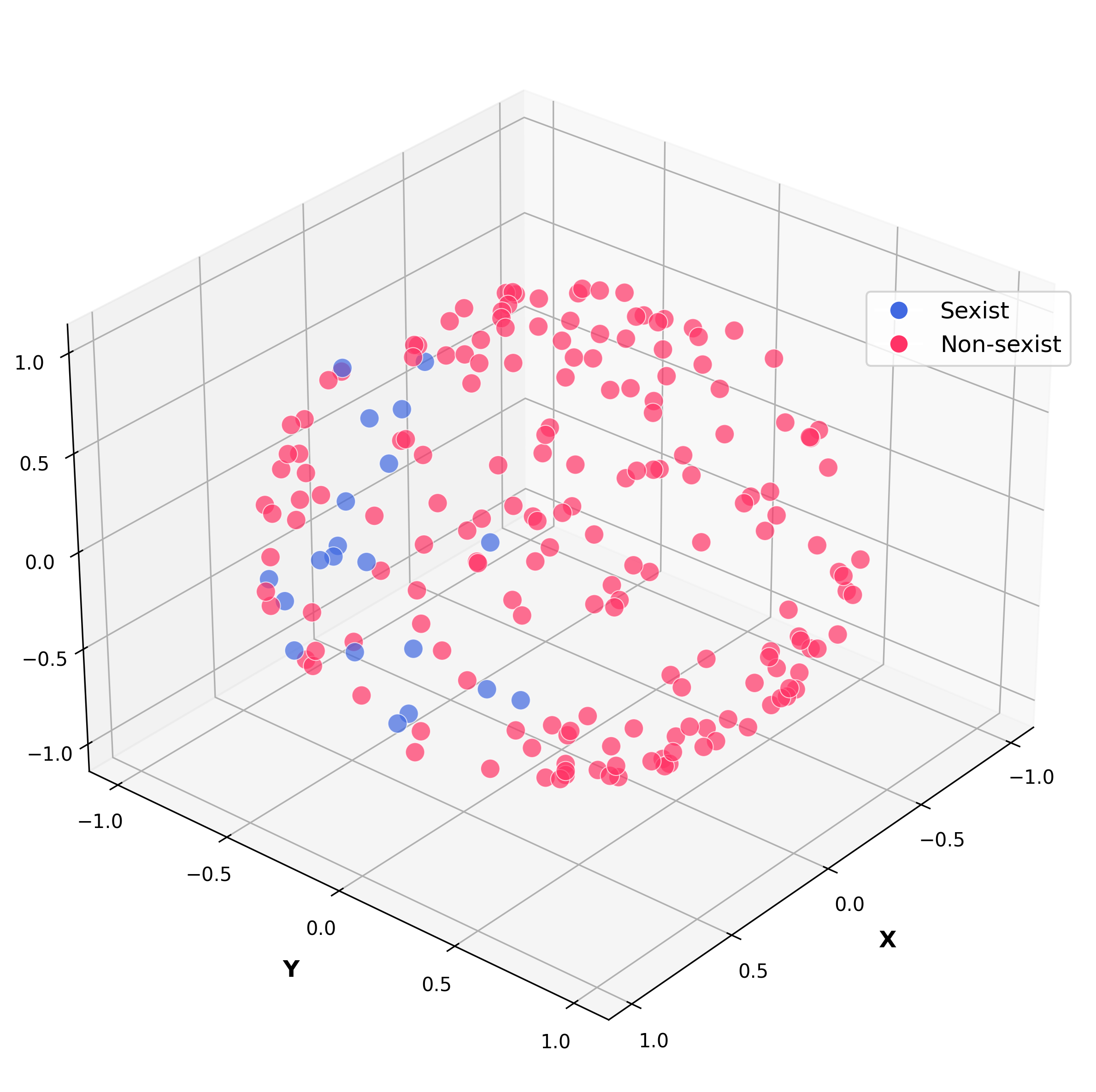}
}
\caption{The minority target class representing deviant language (\bluecircle) versus a highly diverse and heterogeneous non-target class of \textit{everything else} (\redcircle). This t-SNE \cite{maaten2008tsne} visualization (where lighter shades indicate instances located further away in 3-D) displays a representative sample from the ``Call me sexist but \ldots'' corpus \cite{samory2021sexism}.}
\label{fig:distillation-intro}
\vspace{-11pt}
\end{figure}

Most prior research (see \S\hspace{1pt}\ref{sec:related-work}) has approached these tasks as traditional binary classification problems, utilizing ground truth labels provided in various datasets. Despite varying degrees of success, this formulation has overlooked a crucial commonality: the objective is to isolate a minority target class, characterized more by its pragmatic function in natural language than its semantics, from the much larger and incredibly diverse negative class encompassing \textit{everything else}. The complexities arising from the somewhat nebulous dichotomy,\footnote{There is a sizeable body of legal and linguistic scholarship on the boundaries of unwarranted and figurative language. See, for example, \citet{rosenfeld2002hate, kiska2012hate, kasparian2013hemispheric, athanasiadou2024margins}.} alongside the diverse and heterogeneous nature of the predominant non-target class, can be gleaned from~\autoref{fig:distillation-intro}. In this context, accurate binary classification proves challenging due to the immense linguistic diversity encompassed within the non-target category.\footnote{The target class exhibits rich syntactic and semantic variations while serving a specific pragmatic function, whereas the non-target class presents even greater semantic variety with no common pragmatic function. \textit{E.g.}, instances of sexism specifically discriminate on the basis of sex, while the non-target class is unified only by the absence of such hostility.} The model must be adept at learning a complex decision boundary without succumbing to overfitting on the training corpus. Thus, the most effective solutions employ sophisticated deep neural models or ensemble methods, which
\begin{itemize*}
\item[(a)] demand large amounts of training data,
\item[(b)] require significant resources for training and inference,
\item[(c)] lack interpretability, and
\item[(d)] depend on careful regularization and hyperparameter tuning.
\end{itemize*}

\begin{table}[!t]
\setlength{\tabcolsep}{3pt}
\centering
\small
\begin{tabularx}{\linewidth}{@{}c l X@{}}
\toprule
\multirow{5}{*}{\rotatebox[origin=c]{90}{\parbox[c]{120pt}{\centering Figurative}}}
 & $f_1$) & My alarm clock makes sure I love every Monday morning!\hspace{6pt}\greenpatterncircle \\
 & $f_2$) & My alarm clock makes sure that I dread every Monday morning!\hspace{6pt}\redcircle \\
 & $f_3$) & My alarm clock always wakes me up on Monday mornings.\hspace{6pt}\redcircle \\
 & $f_4$) & A white carpet is a great choice when you have messy kids.My alarm clock always wakes me up on Monday mornings.\hspace{6pt}\greenpatterncircle \\
 & $f_5$) & A white carpet is an engaging choice when you have messy kids, if you take extra care.My alarm clock always wakes me up on Monday mornings.\hspace{6pt}\redcircle \smallskip\\
\hdashline[1pt/3pt]\noalign{\vskip 3pt}
\end{tabularx}
\begin{tabularx}{\linewidth}{X X X X}
\;\;\;\;$f_1 \overset{\mathrm{0.47}}{\sim} f_2$ & $f_1 \overset{\mathrm{0.91}}{\sim} f_3$ & $f_2 \overset{\mathrm{0.62}}{\sim} f_3$ & $f_4 \overset{\mathrm{0.78}}{\sim} f_5$ \medskip\\
\end{tabularx}
\begin{tabularx}{\linewidth}{@{}c l X@{}}
\toprule
\multirow{2}{*}{\rotatebox[origin=c]{90}{\parbox[c]{32pt}{\centering Deviant}}}
 & $d_1$) & A female astronaut, because they need sandwiches up there.\hspace{6pt}\greenpatterncircle \\
 & $d_2$) & An astronaut needs sandwiches up there.\hspace{6pt}\redcircle \smallskip\\
\hdashline[1pt/3pt]\noalign{\vskip 3pt}
\end{tabularx}
\begin{tabularx}{\linewidth}{>{\centering\arraybackslash}X}
$d_1 \overset{\mathrm{0.76}}{\sim} d_2$\\
\bottomrule
\end{tabularx}\vspace{-6pt}
\caption{Instances of figurative ($f_i$, sarcasm detection) and deviant ($d_i$, sexism detection) language. Similarity scores are based on the \texttt{stsb-roberta-large} cross-encoder model fine-tuned on the STS benchmark introduced by \citet{cer2017semeval}. These scores reveal a deeper problem: a target class instance (\greenpatterncircle) may be highly \textit{dis}-similar to several non-target instances (\redcircle) while also being very similar to other non-target instances.}
\vspace{-15pt}
\label{tab:clad-whiteboard}
\end{table}

Since the target class has a well-defined function in terms of natural language pragmatics, we conjecture that a detailed study of the target and non-target class distributions will reveal structural differences that can be leveraged to design a model training paradigm better suited for \textbf{cla}ss \textbf{d}istillation (ClaD, discussed in \S\hspace{1pt}\ref{sec:clad}), and therefore, superior in terms of
\begin{itemize*}
\item[(a)] inferring test instances of the target class,
\item[(b)] the demands it places on computational resources during training, and
\item[(c)] interpretability.
\end{itemize*}
We test this conjecture on multiple tasks and benchmark datasets (\S\hspace{1pt}\ref{sec:tasks-datasets}), starting with statistical analyses to glean the \textbf{structural properties and distributional differences between the target and non-target classes} (\S\hspace{1pt}\ref{ssec:structure-distribution}). With these insights, \textbf{we develop novel contrastive loss functions} derived from \citet{mahalanobis1936} distance (\S\hspace{1pt}\ref{ssec:mahalanobis-loss}), which leverage intra-class covariance to contrast the target class against the diverse and heterogeneous collection of negative samples. We then introduce an \textbf{interpretable decision algorithm} based on the normalized squared Mahalanobis distance (\S\hspace{1pt}\ref{ssec:inference}) to identify target instances.

Our results (\S\hspace{1pt}\ref{sec:experiments-results}) demonstrate superior inference and resource efficiency across all tasks. We raise two vital questions in \S\hspace{1pt}\ref{ssec:comparisons-with-llms}, investigating how small language models with ClaD, given limited task-specific training, compares to LLMs in low-resource transfer learning. We also examine the extent to which increasing LLM size improves performance with identical training data. Recent findings, such as those from DeepSeek~\cite{deepseek_v3, deepseek_r1}, underscore the need for such emphasis on economical training and efficient inference.
Further, we present ablation experiments to discern
\begin{itemize*}
\item[(1)] the impact of our decision algorithm, 
\item[(2)] the effect of our novel loss function, and 
\item[(3)] whether traditional one-class classification is comparable to ClaD with Mahalanobis contrast.
\end{itemize*}


\section{ClaD: A Whiteboard Discussion}
\label{sec:clad}
\textbf{Cla}ss \textbf{D}istillation (ClaD) is a specialized training paradigm for binary classification, emphasizing the separation of a distinct category from a diverse and often disproportionately larger non-target background. Non-target instances frequently include expressions that are semantically similar to the target class, while also encompassing elements with no syntactic, semantic, or pragmatic resemblance to each other. This dual challenge leads to significant ambiguity and overlap, making accurate classification particularly difficult.

The predicament is not specific to a single task, as \autoref{tab:clad-whiteboard} shows with instances from figurative (sarcasm) and deviant (sexism) language use. These examples highlight the limitations of relying solely on simple prompting with large language models for effective inference, as decisions are easily confounded by the diversity of the non-target class. Further, we show in \S\hspace{1pt}\ref{ssec:ablation} that applying straightforward semantic similarity measures fails to capture the nuanced characteristics defining the target class.


Drawing insight from \autoref{fig:distillation-intro}, the target instances are not uniformly distributed across the feature space; rather, they appear to form a structured subset that can be viewed as a manifold within a higher-dimensional space. To better understand the shape and properties of this manifold, we analyze the structural and distributional characteristics of the target and non-target classes, providing foundational insights into the class geometries and informing novel loss function formulations (\S\hspace{1pt}\ref{ssec:mahalanobis-loss}) and ClaD's decision algorithm (\S\hspace{1pt}\ref{ssec:inference}).

A visual approach to unveiling the distributional characteristics is relegated to \autoref{app:goodness-of-fit}, while our systematic analysis of the datasets and target class' geometric properties is presented next.
\section{Tasks and Datasets}
\label{sec:tasks-datasets}
We concentrate on three tasks for our analyses and experiments: two types of figurative language (sarcasm and metaphors) and one form of deviant language (sexism), utilizing a dedicated benchmark corpus for each to illustrate that the patterns we uncover and the class distillation paradigm we propose are broadly applicable across such tasks.

\noindent\textbf{1. Sarcasm Headlines (SH)} is a curated dataset comprising professionally crafted headlines from The Onion and HuffPost~\cite{misra2019sarcasm, misra2023Sarcasm}. Notably free of spelling errors and informal language, it offers high-quality labels and self-contained headlines. Compared to social media datasets, it is a clean and reliable resource that precludes concerns about spurious data correlations arising from viral social media trends~\cite{gururangan2018annotation, bender2021parrots}.

\noindent\textbf{2. Trope Finder (TroFi)}~\cite{birke2006clustering} is built to distinguish between literal and non-literal verb usage. It leverages the '88-'89 Wall Street Journal (WSJ) Corpus and enhances it with WordNet, databases of idioms and metaphors, and tags from advanced taggers. TroFi improves metaphor detection by minimizing unverified literal uses and addressing the scarcity of non-literal instances.

\noindent\textbf{3. Call Me Sexist But \ldots (CMSB)} is an innovative corpus designed to detect sexism, comprising tweets that explicitly use the titular phrase to voice potential sexism~\cite{samory2021sexism}. 
It is enhanced with synthetic adversarial modifications to challenge machine learning models. 

\begingroup
\renewcommand{\arraystretch}{1.25}
\setlength{\tabcolsep}{3pt}
\begin{table}[!t]
\small
\centering
\begin{tabularx}{\linewidth}{@{}l l
    >{\raggedleft\arraybackslash}X
    c
    >{\raggedleft\arraybackslash}X
    >{\raggedleft\arraybackslash}X
    >{\raggedleft\arraybackslash}X}
\toprule
Model & Class & \multicolumn{5}{c}{Empirical normality test statistics}\\
\midrule
  & & HZ & & \multicolumn{3}{c}{Anderson-Darling} \\
  & & & & $d_1$ & $d_2$ & $d_3$ \smallskip\\
\multicolumn{7}{@{}l}{Task\,+\,Corpus: Metaphor detection on \textit{TroFi}} \\
\multirow{2}{*}{\textsc{bert} \hspace{6pt} {\LARGE$\lbrace$}}
  & \textit{Metaphor} & 5.14 & & 1.53& 2.59 & 1.96 \\
  & \textit{Other} & 5.93 & & 1.83& 1.71 & 3.40 \\
\multirow{2}{*}{\textsc{s}im\textsc{cse} {\LARGE$\lbrace$}} 
  & \textit{Metaphor} & 4.32 & & 2.83& 2.11 & 0.70 \\
  & \textit{Other} & 5.19& & 2.94 & 2.79 & 1.54 \smallskip\\
\multicolumn{7}{@{}l}{Task\,+\,Corpus: Sarcasm detection on \textit{Sarcasm Headlines}} \\
\multirow{2}{*}{\textsc{bert} \hspace{6pt} {\LARGE$\lbrace$}}
  & \textit{Sarcasm} & 29.94 & & 13.67 & 10.14 & 24.99 \\
  & \textit{Other} & 33.24 & & 13.87 & 23.95 & 5.76 \\
\multirow{2}{*}{\textsc{s}im\textsc{cse} {\LARGE$\lbrace$}}
  & \textit{Sarcasm} & 21.49 & & 19.00 & 16.38 & 23.68 \\
  & \textit{Other} & 24.82 & & 12.21 & 32.65 & 42.99\smallskip\\
\multicolumn{7}{@{}l}{Task\,+\,Corpus: Sexism detection on \textit{Call Me Sexist But}} \\
\multirow{2}{*}{\textsc{bert} \hspace{6pt} {\LARGE$\lbrace$}}
  & \textit{Sexism} & 7.48 & & 0.97 & 1.79 & 9.27 \\
  & \textit{Other} & 34.08 & & 6.70 & 40.34 & 23.96 \\
\multirow{2}{*}{\textsc{s}im\textsc{cse} {\LARGE$\lbrace$}}
  & \textit{Sexism} & 6.38 & & 2.34 & 2.99& 2.08 \\
  & \textit{Other} & 21.21 & & 17.07 & 1.77 & 9.11\\
\bottomrule
\end{tabularx}\vspace{-6pt}
\caption{Empirical results on how well the target and non-target classes fit (a) multivariate normality, using the Henze-Zirkler (HZ) statistic, and (b) univariate normality on the three t-SNE dimensions $d_1, d_2, d_3$, using the Anderson-Darling statistic. For both tests, larger numbers indicate greater deviation from normality.\vspace{-11pt}}
\label{tab:normality-stats}
\end{table}
\endgroup
\subsection{Statistical Tests of Normality}
\label{ssec:structure-distribution}
To systematically analyze the geometric properties of target class representations, we evaluate the normality of the embedding distributions in reduced dimensionality space. Specifically, we apply the HZ~\cite{Henze1990ACO} test
to assess multivariate normality across three t-SNE dimensions for BERT and SimCSE,\footnote{We analyze distributional properties (not downstream performance) using BERT and SimCSE as foundational bidirectional Transformer and contrastive sentence embedding models, respectively. Their selection aligns with established probing protocols prioritizing consistency and transferability across tasks~\cite{reimers2019sentence, rogers2021primer, gao2021simcse}. Findings generalize to architectures like ALBERT and DistilBERT, while rare outliers (GPT-2 and Phi) reflect pretraining misalignment~\cite{ethayarajh2019contextual}, rather than methodological drawbacks.}
examining three pragmatic language detection tasks: metaphor, sarcasm, and sexism. We complement this with AD~\cite{Anderson1952AsymptoticTO} tests
to evaluate univariate normality along individual dimensions. The results (Table~\ref{tab:normality-stats}) offer a rigorous statistical characterization of the manifold structure, beyond the Q-Q plot inspections shown in \autoref{app:goodness-of-fit}.

The HZ tests consistently show lower values for target data (metaphor, sarcasm, and sexism) compared to non-target data, indicating that target data are closer to a multivariate normal distribution compared to their non-target counterparts. This is further supported by the results of the AD tests along each dimension. Reduced deviation from the theoretical distribution suggests that the target data exhibits a more homogeneous manifold structure. In contrast, the non-target data manifests greater diversity and complexity. 
Thus, in line with the argument presented earlier with illustrative examples (\S\hspace{1pt}\ref{sec:clad}), it is indeed less likely that they possess discernible common traits beyond their opposition to the target class. Hence, we hypothesize that a loss function ought to be designed primarily around the target class. The consistency and regularity of the target class' distribution provide a more reliable foundation for learning stable predictors, in contrast to the somewhat more chaotic diversity observed in the distribution of the non-target class.

\section{Training and Inference}
\label{sec:training-inference}
\citet{mahalanobis1936} distance is ideal for data approximating a multivariate normal distribution, as it accounts for the manifold structure of the target class, rendering the distance measure scale-invariant. By incorporating the variance and correlations among variables, it accurately reflects the underlying distribution of the data and thereby improves discrimination in detecting non-target instances by robust identification of outliers, reducing false positives.
Accordingly, we explore Mahalanobis distance in formulating the loss function.


\subsection{Mahalanobis Loss}
\label{ssec:mahalanobis-loss}
Let $\mathcal{X} = \{x_i\}, \mathcal{Y} = \{y_j\}$ denote $n$ target and $m$ non-target training samples (resp.). Further, let $f: \mathcal{X} \cup \mathcal{Y} \mapsto \mathbb{R}^d$ denote a representation function mapping these instances to $d$ dimensions. For a given instance $x \in \mathcal{X}$, we randomly select 
$x^+ \in \mathcal{X} \setminus \{x\}$, and 
$y^- \in \mathcal{Y}$. These random selections are employed to learn a representation that minimizes (maximizes) the similarity between $x$ and $y^-$ ($x^+$). 
We achieve this with \textbf{Mahalanobis loss}:\vspace*{-14pt}

{\small
\begin{equation}
\mathcal{L}_{\textsc{mah}} =\frac{1}{|\mathcal{X}|} \sum_{x\in \mathcal{X}} \frac{\text{sim}_{\textsc{mah}}(x, y^-)}{\text{sim}_{\textsc{mah}}(x, x^+) + \text{sim}_{\textsc{mah}}(x, y^-)}
\label{eq:mah_loss}
\end{equation}
}%
where $\text{sim}_{\textsc{mah}}(x, y)$ is defined using the covariance matrix $\Sigma$ of the set $\{f(x_i)\}$
\vspace{-16pt}

{\small
\begin{multline*}
\text{sim}_{\textsc{mah}}(x, y) = \\
\exp\left\{- \frac{(f(x) - f(y))^T \Sigma^{-1} (f(x) - f(y))}{d}\right\}.
\end{multline*}
}%
\vspace{-16pt}

Alternatively, \textbf{Mahalanobis mean loss} uses the mean $\mu$ of $\{f(x_i)\}$:\vspace*{-11pt}

{\small
\begin{equation}
\begin{aligned}
\mathcal{L}_{\textsc{mah},\mu}
&=
-\frac{1}{|\mathcal{X}|}
\sum_{(x,y^-)\in \mathcal{X}}
\Bigl[
\,\log\bigl(\text{sim}_{\textsc{mah}}(\mu, x)\bigr)
\\
&\quad
+\;
\log\bigl(1 - \text{sim}_{\textsc{mah}}(\mu, y^-)\bigr)
\Bigr].
\end{aligned}
\label{eq:mah_mean_loss}
\end{equation}
}%
It maximizes the similarity between a target instance $x$ and the mean representation of the target class (making the class more compact), and minimizes the similarity between a negative example $y^-$ and the mean (increasing inter-class margin).

\begin{algorithm}[!t]
\begin{algorithmic}
{\small
\Require New instance \(X = X^*\)
\State \(X_{n+1} \gets X^*\)
\State \(\hat{\mu} \gets \frac{1}{n+1} \sum_{i=1}^{n+1} X_i\)
\State Compute \(\hat{\Sigma}\) as the sample covariance matrix of \(X_1, \dots, X_{n+1}\)
\State \(d_{n+1}^2(\hat{\mu}, \hat{\Sigma}) \gets (X_{n+1} - \hat{\mu})^T \hat{\Sigma}^{-1} (X_{n+1} - \hat{\mu})\)
\State \(T \gets \frac{n+1}{n^2} d_{n+1}^2(\hat{\mu}, \hat{\Sigma})\)
\State Compute the critical value $v_\beta$ for {\small Beta\(\left(\frac{d}{2}, \frac{n-d}{2}\right)\)}
\If{$T < v_\beta$}
\State \(X \gets 1\) \Comment{Target class}
\Else
\State \(X \gets 0\) \Comment{Non-target class}
\EndIf
}
\end{algorithmic}
\caption{Mahalanobis $\beta$-decision algorithm} 
\label{alg:decision}
\end{algorithm}

\begin{figure*}[!t]
\begin{subfigure}[t]{\textwidth}
    \centering
    \begin{minipage}{0.25\textwidth}
        \centering
        \includegraphics[width=\linewidth]{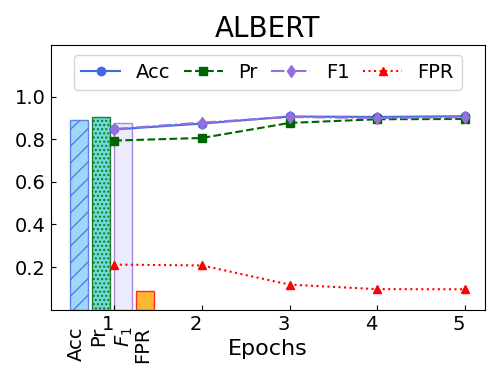}
    \end{minipage}%
    \hfill
    \begin{minipage}{0.25\textwidth}
        \centering
        \includegraphics[width=\linewidth]{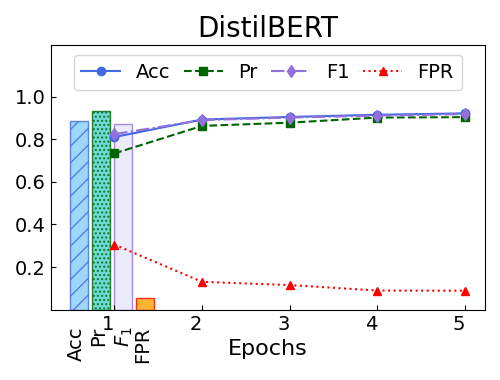}
    \end{minipage}%
    \hfill
    \begin{minipage}{0.25\textwidth}
        \centering
        \includegraphics[width=\linewidth]{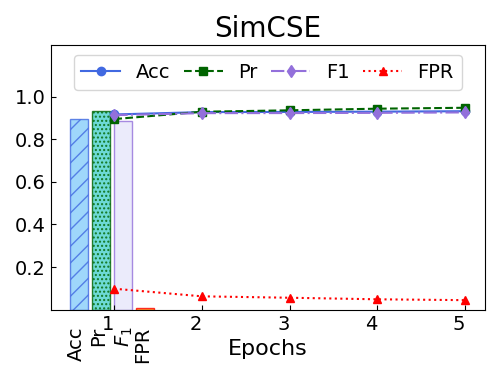}
    \end{minipage}%
    \hfill
    \begin{minipage}{0.25\textwidth}
        \centering
        \includegraphics[width=\linewidth]{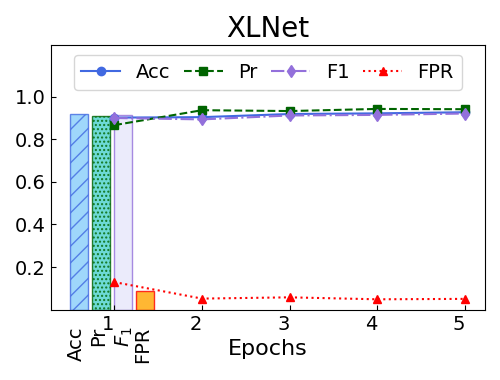}
    \end{minipage}
    \caption{Figurative language: sarcasm detection using the \textit{Sarcasm Headlines} (SH) corpus~\cite{ misra2023Sarcasm}.}
    \label{fig:row1}
\end{subfigure}%

\vspace{1em}

\begin{subfigure}[t]{\textwidth}
    \centering
    \begin{minipage}{0.25\textwidth}
        \centering
        \includegraphics[width=\linewidth]{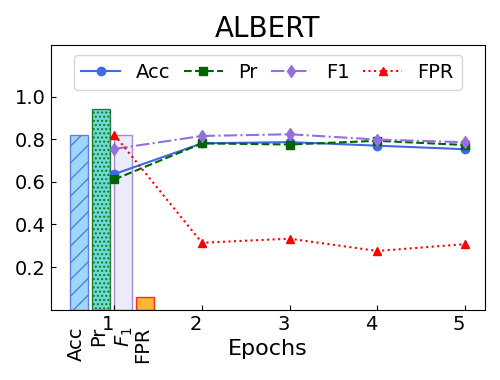}
    \end{minipage}%
    \hfill
    \begin{minipage}{0.25\textwidth}
        \centering
        \includegraphics[width=\linewidth]{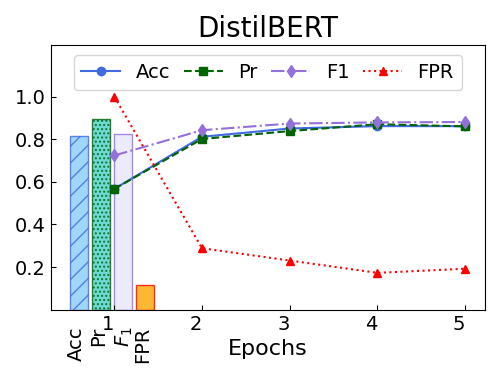}
    \end{minipage}%
    \hfill
    \begin{minipage}{0.25\textwidth}
        \centering
        \includegraphics[width=\linewidth]{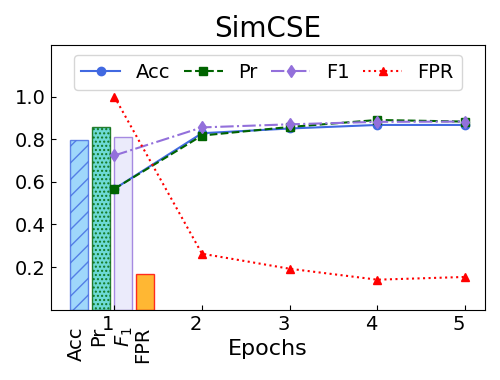}
    \end{minipage}%
    \hfill
    \begin{minipage}{0.25\textwidth}
        \centering
        \includegraphics[width=\linewidth]{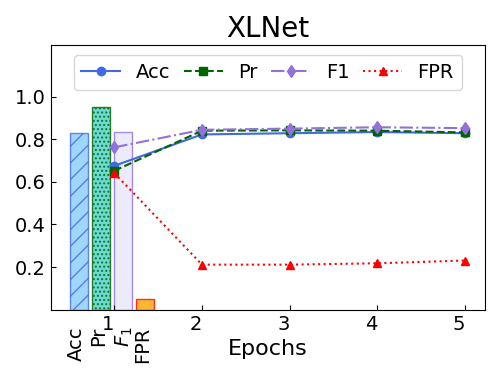}
    \end{minipage}
    \caption{Figurative language: metaphor detection using the \textit{Trope Finder} (TroFi) corpus~\cite{birke2006clustering}.}
\end{subfigure}%

\vspace{1em}

\begin{subfigure}[t]{\textwidth}
    \centering
    \begin{minipage}{0.25\textwidth}
        \centering
        \includegraphics[width=\linewidth]{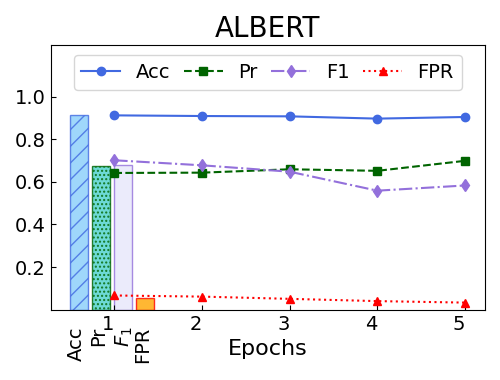}
    \end{minipage}%
    \hfill
    \begin{minipage}{0.25\textwidth}
        \centering
        \includegraphics[width=\linewidth]{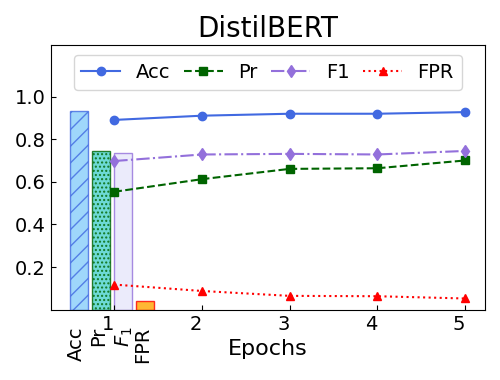}
    \end{minipage}%
    \hfill
    \begin{minipage}{0.25\textwidth}
        \centering
        \includegraphics[width=\linewidth]{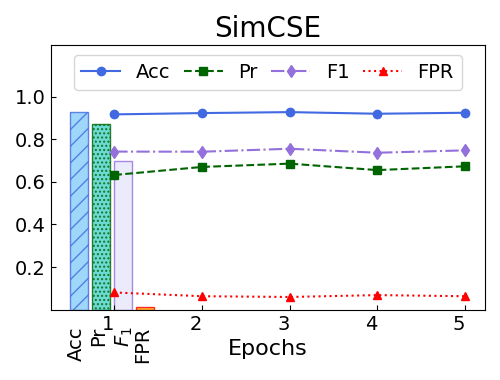}
    \end{minipage}%
    \hfill
    \begin{minipage}{0.25\textwidth}
        \centering
        \includegraphics[width=\linewidth]{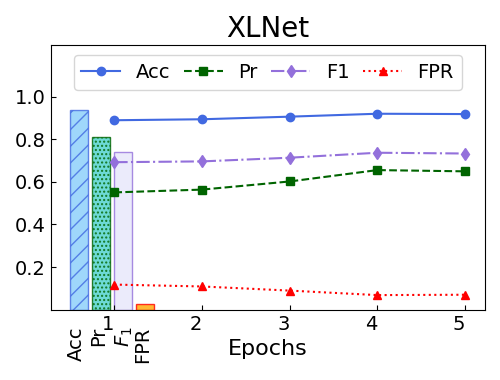}
    \end{minipage}
    \caption{Deviant language: sexism detection using the \textit{Call me sexist but \ldots} (CMSB) corpus~\cite{samory2021sexism}.}
\end{subfigure}%
\caption{Comparison of ClaD across three detection tasks (from top to bottom) -- (a) sarcasm, (b) metaphors, and (c) sexism -- against four transfer learning baseline results where Transformer-based models are fine-tuned on task-specific data: (from left to right) ALBERT, DistilBERT, SimCSE, and XLNet.}
\label{fig:transfer-learning-baselines}
\vspace{-11pt}
\end{figure*}
\subsection{Inference and Decision Algorithm}
\label{ssec:inference}
The inference task is, fundamentally, identical to that of any supervised binary classifier: ascertain if a test instance belongs to the target class. Given representations $\{\mathbf{x}_1, \mathbf{x}_2, \ldots, \mathbf{x}_n\}$ with sample mean $\hat{\mathbf{\mu}}$ and covariance matrix $\Sigma$ adhering to a multivariate normal distribution, the squared Mahalanobis distance for a specific observation $\mathbf{x}_i$ is given by\vspace*{-11pt}

{\small
\begin{equation}
d_i^2(\hat{\mathbf{\mu}}, \Sigma) =
  (\mathbf{x}_i - \hat{\mathbf{\mu}})^T\Sigma^{-1}(\mathbf{x}_i - \hat{\mathbf{\mu}}),
  \label{eq:sq-mah_distance}
\end{equation}
}%
which follows the Beta distribution~\cite{wilks1962mathematical, ververidis2008gaussian}:\vspace*{-11pt}

{\small
\begin{equation}
\frac{n}{(n-1)^2}\,d_i^2(\hat{\mathbf{\mu}}, \Sigma) \sim \textrm{Beta}\left(\frac{d}{2},\, \frac{n-d-1}{2}\right)   
\end{equation}
}%
This insight informs the design of the \textbf{Mahalanobis $\boldsymbol{\beta}$-decision algorithm} (Algorithm~\ref{alg:decision}), to test an instance for class membership by comparing its normalized squared Mahalanobis distance to critical values of the corresponding Beta distribution.\footnote{The critical threshold value is determined based on development data, ensuring optimal calibration for inference.}
\section{Experiments and Results}
\label{sec:experiments-results}
We empirically evaluate\footnote{\autoref{app:config} contains implementation details (\S\hspace{1pt}\ref{ssec:comparisons-with-old-models}-\S\hspace{1pt}\ref{ssec:comparisons-with-llms}).} ClaD across two challenging categories of language understanding tasks: detecting figurative (metaphor and sarcasm) and harmful (sexism) language. ClaD leverages the Mahalanobis mean loss, $\mathcal{L}_{\textsc{mah},\mu}$ (Eq.~\ref{eq:mah_mean_loss}), to fine-tune pretrained embeddings, followed by inference with the Mahalanobis $\beta$-decision algorithm (Alg.~\ref{alg:decision}).

ClaD is benchmarked against two modern language model paradigms:
\begin{enumerate*}[label=(\alph*)]
\item specialized encoder(-decoder) architectures optimized for language understanding: SimCSE~\cite{gao2021simcse}, ALBERT~\cite{lan2020albert}, DistilBERT~\cite{sanh2019distilbert}, and XLNet~\cite{yang2019xlnet},\footnote{DeBERTa~\cite{he2021deberta} 
performs much like ALBERT. So, for architectural diversity, we include XLNet instead.} and
\item large language models (LLMs), primarily decoder-only architectures distinguished by their scale.
\end{enumerate*}
We evaluate all models on 80/10/10 splits for train/dev/test, primarily focusing on the false positive rate (FPR) and $F_1$ score for the target class.\footnote{Reducing false positives in these tasks is particularly important in several applications such as social media moderation, so as to not penalize users for innocuous remarks}

\begin{figure*}
\centering
\begin{subfigure}[t]{0.33\textwidth}
  \centering
  \includegraphics[width=0.98\textwidth]{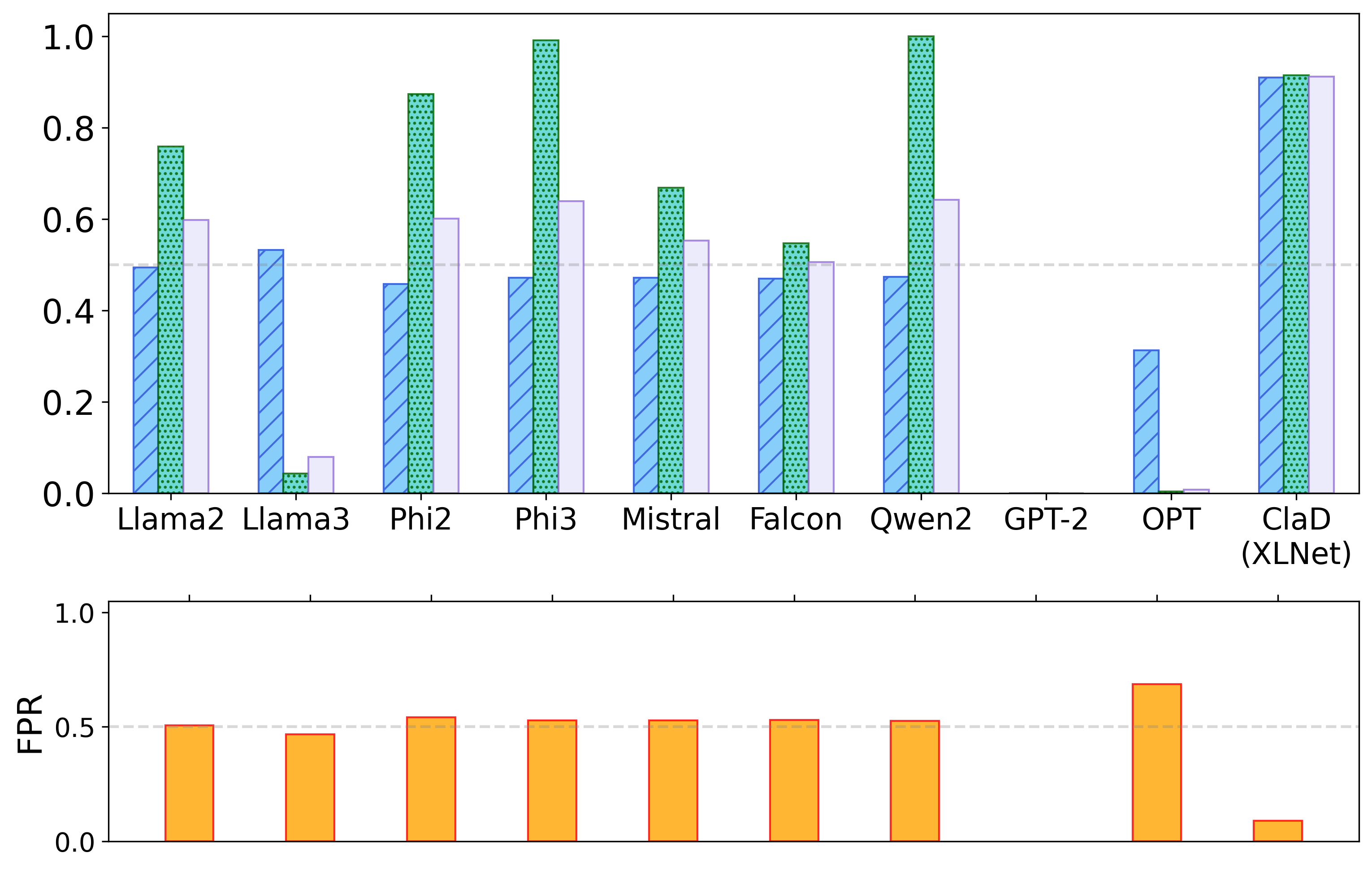}
  \caption{Figurative language: sarcasm}
  \end{subfigure}%
    ~ 
\begin{subfigure}[t]{0.33\textwidth}
  \centering
  \includegraphics[width=0.98\textwidth]{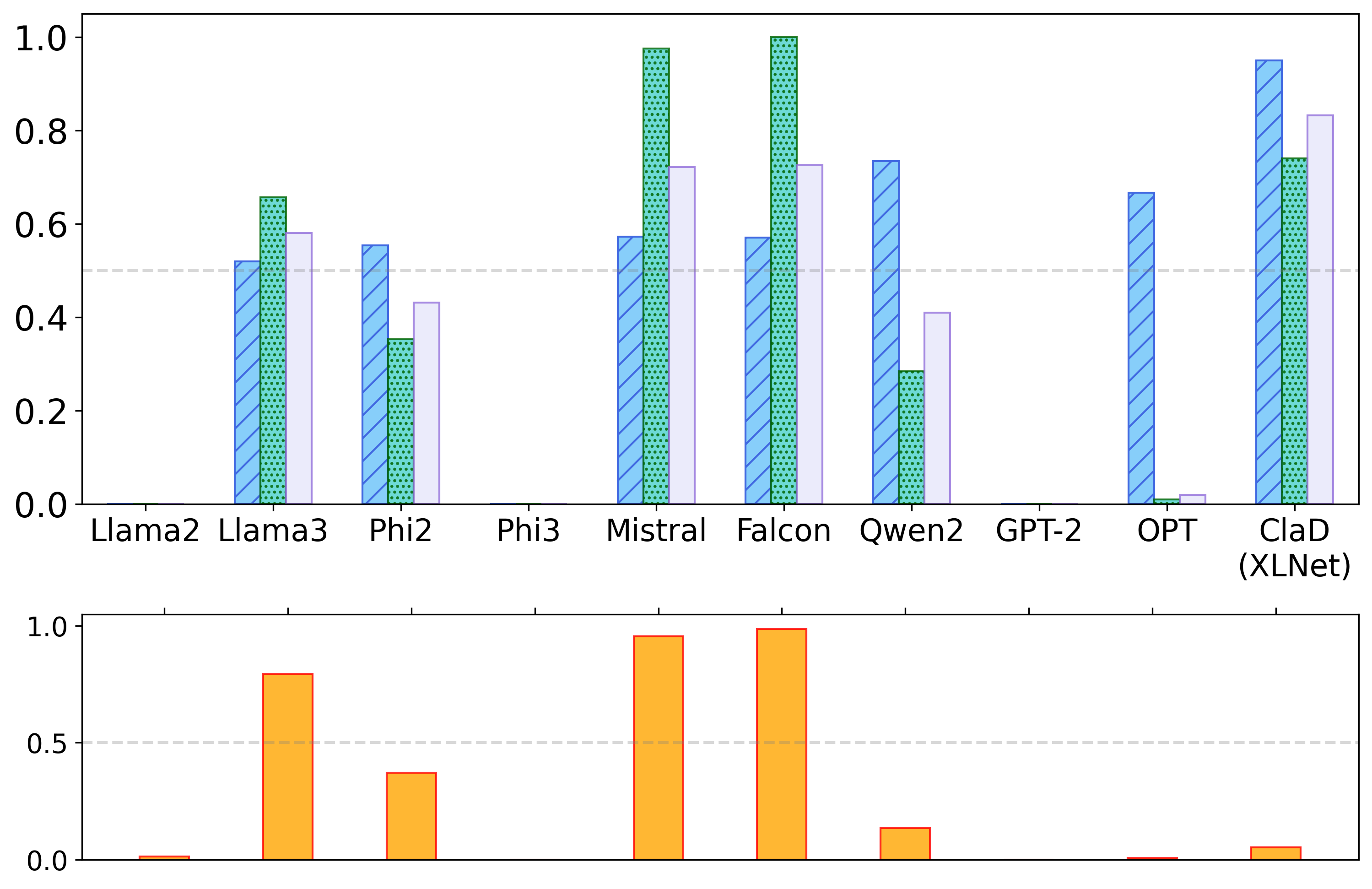}
  \caption{Figurative language: metaphors}
\end{subfigure}%
    ~ 
\begin{subfigure}[t]{0.33\textwidth}
  \centering
  \includegraphics[width=0.98\textwidth]{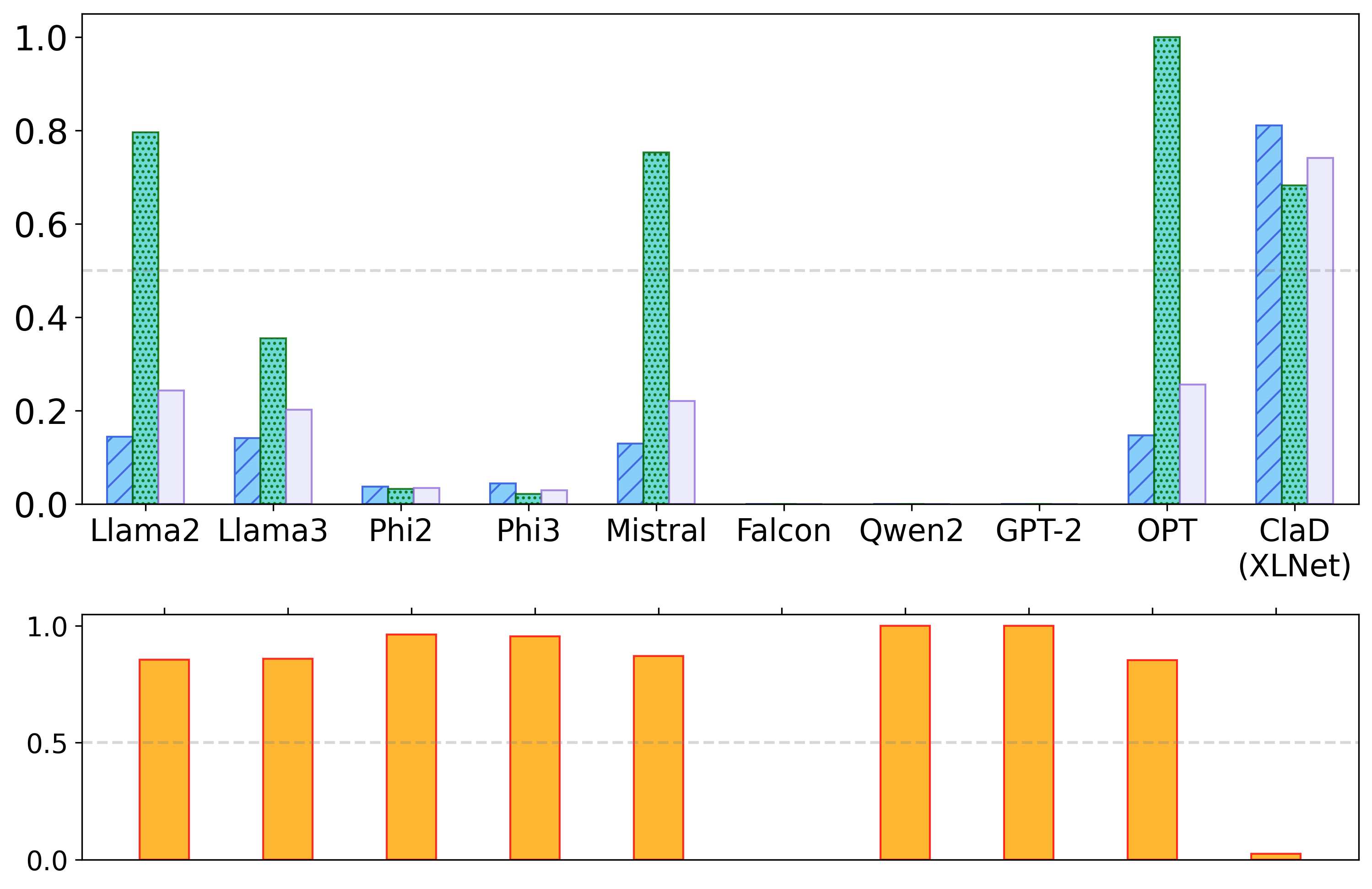}
  \caption{Deviant language: sexism}
\end{subfigure}
\caption{Comparison of 5-shot evaluation of a suite of nine large language models (left to right): Llama2, Llama3, Phi2, Phi3, Mistral-7B, Falcon, Qwen2, GPT-2, and OPT, against ClaD's single-epoch training (rightmost).}
\label{fig:five-shot-llm-results}
\vspace{-11pt}
\end{figure*}
\subsection{Comparison with Encoder-based Models}
\label{ssec:comparisons-with-old-models}~\autoref{fig:transfer-learning-baselines} illustrates ClaD's distinctive advantage: achieving performance competitive with or superior to these established models in just one training epoch. In contrast, transfer learning with these models typically requires 3-5 epochs to attain similar metrics across all three tasks. ClaD's rapid convergence 
yields substantial computational savings without compromising detection quality. For instance, in sarcasm detection, ClaD achieves a lower FPR after one epoch than most baselines do after five.\footnote{Only XLNet marginally surpasses ClaD after epoch 2, with a difference of 0.036. A bootstrap analysis reveals this as statistically insignificant: ClaD's FPR (8.87\%) is well within the 95\% CI (5.05\%, 9.93\%) of XLNet's mean FPR.}
Similar patterns emerge across all tasks and metrics, where ClaD's single-epoch training matches or outperforms multi-epoch training of the transfer-learning baseline models. The results suggest that ClaD's geometric approach enables efficient adaptation to task-specific features, a finding further supported by our ablation study (\S\hspace{1pt}\ref{ssec:ablation}).

\subsection{Comparison with Large Language Models}
\label{ssec:comparisons-with-llms}
Next, we evaluate ClaD against a suite of recent large language models (LLMs): OPT, GPT, Phi, Llama, Mistral, Qwen, and Falcon. As ClaD is a training paradigm, and not a model, these evaluations are geared to answer two research questions:

{
\renewenvironment{quote}
  {\list{}{\leftmargin=3mm \rightmargin=3mm \topsep=0pt} \item }
  {\endlist}
\begin{quote}
\textbf{Q1.} \textit{Is limited task-specific ClaD-training with small language models better than low-resource transfer-learning with LLMs?}
\end{quote}
\begin{quote}
\textbf{Q2.} \textit{With identical training data, how much larger are the LLMs (if any) that outperform ClaD-training with small language models?}
\end{quote}
}

Against all LLMs in zero-shot (\autoref{app:zero-shot}) and few-shot scenarios (discussed next), ClaD demonstrates consistently superior performance.

\paragraph{Few-shot Classification:}
As shown in \autoref{fig:five-shot-llm-results}, ClaD retains its substantial advantage over few-shot classification (five instances) with LLMs, which achieve markedly lower $F_1$ scores: from 0.0\% (GPT-2) to 64.2\% (Qwen-2) on \textit{SH}; 0.0\% (GPT-2) to 72.6\% (Falcon) on \textit{TroFi}, and 0.0\% (Falcon) to 25.6\% (OPT) on \textit{CMSB}. 
Models exhibit distinct characteristics in each task: for example, in metaphor detection, GPT-2, Llama2, and Phi3 completely avoid positive predictions, while Falcon predicts aggressively with perfect recall.
The most dramatic changes are seen in deviant language detection, with more models completely avoiding the target class (Falcon, Qwen2, and GPT-2). OPT, on the other hand, exhibits perfect recall. Despite accuracy ranges similar to zero-shot, models show more extreme precision-recall trade-offs, with AUC scores stalled near 0.5 and persistently high FPR.

LLMs thus exhibit notable limitations and variability across all tasks, likely stemming from insufficient feature learning in low-resource scenarios (reflected in the AUC stagnation), causing them to fall back on their pretrained biases, particularly for subtle, context-dependent linguistic cues. The erratic behavior changes between zero- and few-shot settings also suggest unstable optimization paths, possibly due to the large parameter count in these models, stochasticity of gradient updates, and insufficient regularization.

\begin{figure*}[!t]
    \centering
    \begin{subfigure}[t]{0.33\textwidth}
        \centering
        \includegraphics[width=\textwidth]{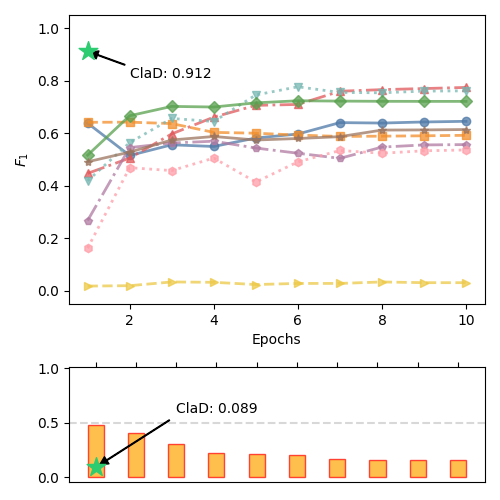}
        \caption{Figurative language: sarcasm}
    \end{subfigure}%
    ~
    \begin{subfigure}[t]{0.33\textwidth}
        \centering
        \includegraphics[width=\textwidth]{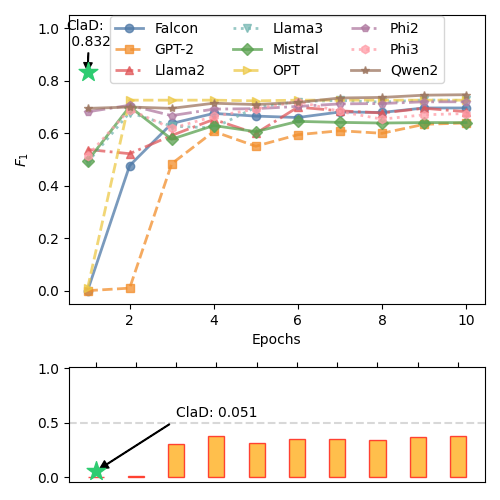}
        \caption{Figurative language: metaphors}
    \end{subfigure}%
    ~
    \begin{subfigure}[t]{0.33\textwidth}
        \centering
        \includegraphics[width=\textwidth]{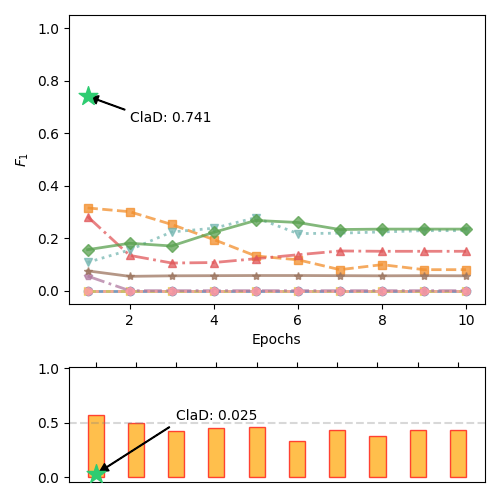}
        \caption{Deviant language: sexism}
    \end{subfigure}
\caption{Comparison of ClaD across the three detection tasks against nine large language models (LLMs) in a limited data regime. The LLMs are trained on 100 instances over 10 epochs. Results shown for the target class are: (top) the $F_1$ scores; and (bottom) the false positive rates (FPR) for the best-performing LLM.}
\label{fig:f1-low-resource}
\end{figure*}

\begin{figure*}[!t]
    \centering
    \begin{subfigure}[t]{0.33\textwidth}
        \centering
        \includegraphics[width=\textwidth]{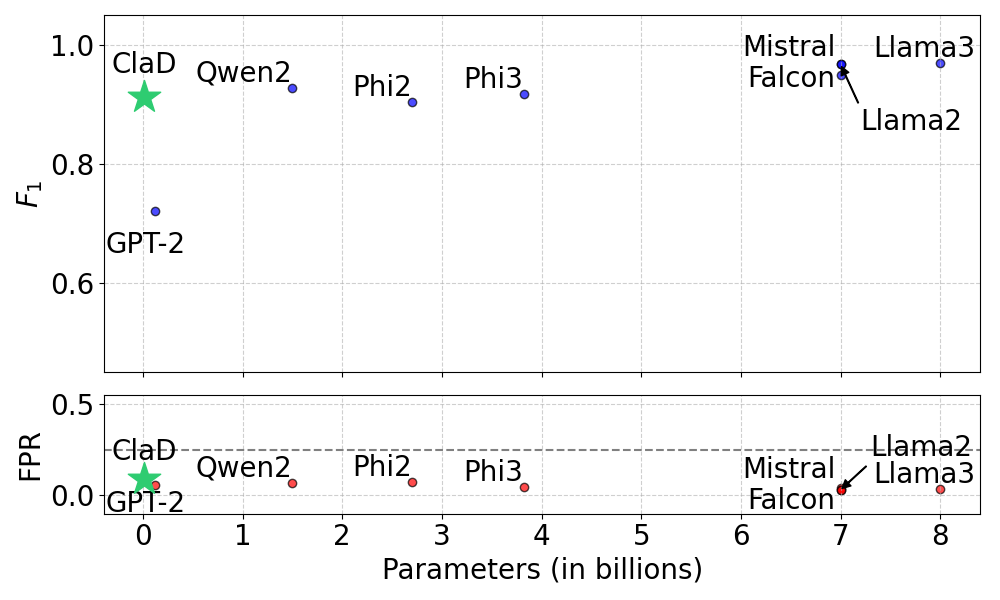}
        \caption{Figurative language: sarcasm}
    \end{subfigure}%
    ~
    \begin{subfigure}[t]{0.33\textwidth}
        \centering
        \includegraphics[width=\textwidth]{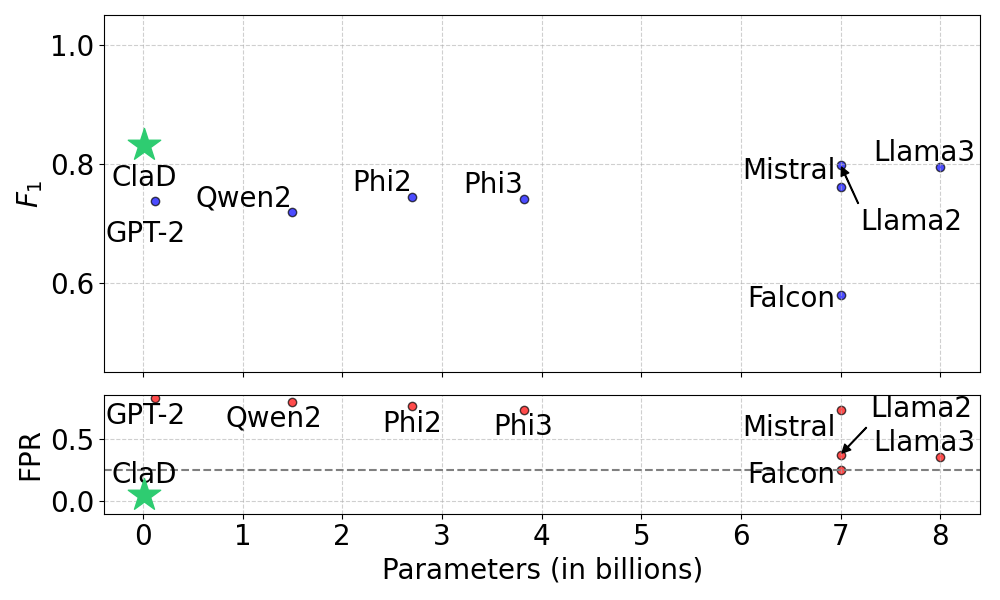}
        \caption{Figurative language: metaphors}
    \end{subfigure}%
    ~
    \begin{subfigure}[t]{0.33\textwidth}
        \centering
        \includegraphics[width=\textwidth]{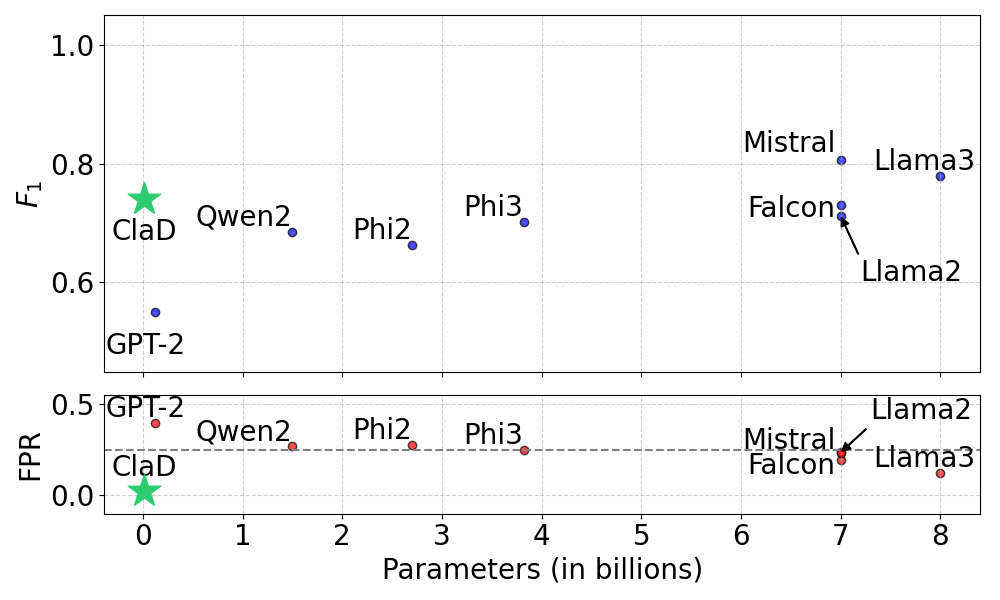}
        \caption{Deviant language: sexism}
    \end{subfigure}
\caption{Comparison of ClaD across the three detection tasks against LLMs, with identical training data: all models utilize the entire training set for a single epoch. $F_1$ scores (\textit{top}) show ClaD being competitive with most LLMs, and outperforming a few others, while false positive rates (FPR) (\textit{bottom}) show ClaD remaining superior to the LLMs.}
\vspace{-11pt}
\label{fig:f1-full-resource}
\end{figure*}

\paragraph{Low-resource Training:}
We extend our analysis to low-resource training (with 100 instances provided to the LLMs) to examine whether this limited increase in data improves decision stability, precision-recall trade-offs, and task adaptation. ClaD's single-epoch training continues to outperform all LLMs in terms of both $F_1$ score and FPR, except in metaphor detection. There, although GPT-2, Falcon, and OPT report lower FPRs for the first two epochs, their $F_1$ scores are nearly zero as they completely avoid false positives (which comes at the cost of failing to avoid \textit{any} positives).
\autoref{fig:f1-low-resource} reveals clear patterns across model scales: while the smaller XLNet-based ClaD achieves superior performance within one epoch, larger models like Llama3 (8B parameters) require multiple epochs to reach their peak performance (\textit{e.g.}, 77.6\% $F_1$ at epoch 6 for sarcasm detection). Model size significantly impacts learning trajectories: large models (7B-8B parameters) show rapid initial improvements, mid-size models (2B-4B parameters) plateau early with suboptimal performance, and smaller models (124M-350M parameters) struggle to learn effectively. Sexism detection remains the most challenging task, with all LLMs showing conservative labeling of the target class and an inability to learn from limited data. In sarcasm detection, the only task where LLMs perform significantly better than chance, FPR correlates inversely with model size, ranging from 15.4\% (Llama3, epoch 8) to 48.1\% (GPT-2, epoch 10).

\paragraph{Identical Task-specific Training Data:} To address our second research question, we compare ClaD's single-epoch training with smaller models against the suite of LLMs (\autoref{fig:f1-full-resource}).\footnote{OPT markedly underperformed across all tasks and evaluation metrics, and is thus excluded in the comparison.}

The relationship between model size and performance varies significantly across tasks. While larger models (7-8B parameters) generally perform well in sarcasm detection ($F_1$: 0.96-0.97), this advantage diminishes in metaphor detection ($F_1$: 0.71-0.81) and almost disappears in sexism detection, where many smaller models
achieve competitive performance. This suggests that larger models do not consistently translate to better performance across pragmatic language understanding tasks.

More often than not, performance improves with increased model size, but with diminishing returns. A striking pattern emerges in the false positive rates, however: while larger models show very low FPR in sarcasm detection ($\sim$ 0.03), their FPR varies widely in other tasks, sometimes performing worse than smaller models. Particularly interesting is deviant language (sexism) detection, where even the largest models struggle with high FPR. The consistent performance of XLNet-based ClaD, with only 110M params, especially in maintaining lower FPR across tasks while remaining competitive on $F_1$ scores, suggests that efficiency derived from a geometric understanding eclipses model size.


\subsection{Ablation Experiments}
\label{ssec:ablation}
We conduct systematic ablation experiments to evaluate the individual contributions of ClaD's core components: the novel loss function, the decision algorithm, and the Mahalanobis contrast mechanism. We present comparisons using SimCSE as the base model, as it achieved the lowest false positive rate in sarcasm detection.\footnote{ClaD (XLNet) reports marginally better $F_1$, and experiments with XLNet as the base model yield similar results.} The impact of our novel loss function is evident in \autoref{tab:ablation-1}. Compared to standard loss functions in task-specific fine-tuning, $F_1$ scores improve by 26.5\%, 6\%, and 13\% for sarcasm, metaphor, and sexism detection, respectively (with corresponding proportionate decreases seen in FPR: 24.6\%, 39.1\%, and 15.3\%). On the other hand, replacing our $\beta$-decision algorithm with a 3-layer fully connected feed-forward network for classification results in the $F_1$ scores dropping by 58.5\%, 80.8\%, and 46.0\% on these tasks, respectively.
Finally, we show in \autoref{app:ablation} that traditional one-class classification and anomaly detection methods do not perform well in these pragmatic language tasks where the minority target class requires gleaning from a large heterogeneous non-target majority.

\begingroup
\renewcommand{\arraystretch}{1.25}
\setlength{\tabcolsep}{3pt}
\begin{table}[!t]
\small
\centering
\begin{tabularx}{\linewidth}{@{}l
    >{\raggedleft\arraybackslash}X
    >{\raggedleft\arraybackslash}X
    >{\raggedleft\arraybackslash}X
    >{\raggedleft\arraybackslash}X}
\toprule
 & Acc \ua{} & Pr \ua{} &  FPR \da{} & $F_1$ \ua{} \\
\midrule
\multicolumn{5}{c}{Sarcasm detection on \textit{Sarcasm Headlines}} \\
$\mathcal{L}_{\textsc{mah}}$ + $\beta$-decision & 0.896 & 0.931 & 0.009 & 0.885 \\
$\mathcal{L}_{\textsc{cosine}}$ + $\beta$-decision & 0.492 & 0.479 & 0.255 & 0.620 \\
$\mathcal{L}_{\textsc{mah}}$ + MLP & 0.521 & 0.482 & 0.209 & 0.300\medskip\\
\multicolumn{5}{c}{Metaphor detection on \textit{TroFi}} \\
$\mathcal{L}_{\textsc{mah}}$ + $\beta$-decision & 0.794 & 0.857 & 0.167 & 0.808 \\ 
$\mathcal{L}_{\textsc{cosine}}$ + $\beta$-decision & 0.675 & 0.667 & 0.558 & 0.748 \\
$\mathcal{L}_{\textsc{mah}}$ + MLP & 0.433 & 0.000  & 0.000 & 0.000\medskip\\
\multicolumn{5}{c}{Sexism detection on \textit{Call Me Sexist But}} \\
$\mathcal{L}_{\textsc{mah}}$ + $\beta$-decision & 0.928 & 0.870 & 0.012 & 0.696 \\
$\mathcal{L}_{\textsc{cosine}}$ + $\beta$-decision &  0.832 & 0.433 & 0.165 & 0.566 \\
$\mathcal{L}_{\textsc{mah}}$ + MLP & 0.134 & 0.134 & 1.000 & 0.236\smallskip\\
\bottomrule
\end{tabularx}
\caption{Ablation study comparing ClaD's components: $\mathcal{L}_{\textsc{mah}}$ (our Mahalanobis contrast loss), $\mathcal{L}_{\textsc{cosine}}$ (the standard cosine similarity loss), and $\beta$-decision (Algorithm~\ref{alg:decision}). The combination of $\mathcal{L}_{\textsc{mah}}$ + $\beta$-decision achieves superior performance across all metrics, particularly in reducing false positive rates (FPR) while maintaining high $F_1$ and target-class precision.}
\vspace{-11pt}
\label{tab:ablation-1}
\end{table}
\endgroup
\section{Related Work}
\label{sec:related-work}
Extensive research aims to model pragmatic language nuances for respectful communication, advancing the identification of figurative language~\cite{chakrabarty2022flute, saakyan2022report, wachowiak2023gpt, huiyuan2024survey} and deviant content~\cite{fortuna2018survey, yin2021towards, guest2021expert, bose2022deep}.
Most leverage BERT-based supervised learning: e.g., BERT-BiLSTM for hate speech~\cite{bose2022deep}, dual BERT models for metaphors~\cite{wan2021enhancing}, and BERT-LSTM for sarcasm~\cite{kumar2020transformers}. Enhancements include syntactic~\cite{wan2020using} or semantic~\cite{zhou2021hate} feature integration and multi-task frameworks~\cite{samghabadi2020aggression}. However, generalization remains limited, and ensemble methods~\cite{lemmens2020sarcasm, gregory2020transformer} trade interpretability for computational cost. Our Class Distillation (ClaD) paradigm addresses these gaps via an interpretable decision algorithm and novel loss function.

ClaD shares similarities with one-class classification, which detects anomalies by focusing on the target class. Common methods include one-class SVM~\cite{DBLP:journals/neco/ScholkopfPSSW01, noumir2012occ}, DeepSVDD~\cite{DBLP:conf/icml/RuffGDSVBMK18}, and adversarial one-class classifiers~\cite{DBLP:conf/cvpr/SabokrouKFA18}, but they often struggle with domain generalization, overfitting, and nuanced data -- key challenges in pragmatic language tasks. ClaD, leveraging Mahalanobis contrast, effectively addresses these issues, demonstrated by ablation results in \autoref{app:ablation}.


LLMs like GPT-2~\cite{radford2019language}, Phi-2~\cite{javaheripi2023phi}, and OPT~\cite{zhang2022opt} excel in text classification but are computationally expensive~\cite{DBLP:journals/corr/abs-2312-01044}. Some, like GPT-3, reportedly struggle with nuanced tasks like metaphor detection~\cite{wachowiak2023gpt}, while others face reasoning limitations and token constraints in in-context learning~\cite{DBLP:conf/emnlp/SunL0WGZ023}. Unlike recent efforts to address these issues, ClaD combines Mahalanobis contrast with \textit{smaller} models, to \textit{efficiently} learning task manifolds.

\section{Conclusion}
\label{sec:conclusion}
This work challenges a fundamental implicit assumption in modern NLP: that scale (in models, pretraining data, or fine-tuning) guarantees superior downstream performance. Through rigorous empirical analysis, we demonstrate that our geometrically grounded training paradigm 
surpasses state-of-the-art LLMs by significant margins in low-data regimes, achieving superior results in a single epoch where larger models plateau after several. Notably, ClaD matches or exceeds the performance of models nearly two orders of magnitude larger, \textit{even with identical task-specific training}.

Our findings align with broader trends toward efficiency, spurred by DeepSeek's compute-optimal scaling~\cite{deepseek_v3, deepseek_r1}, and reveal a novel insight: \textbf{architectural minimalism}, coupled with \textbf{geometric alignment to task manifolds}, can unlock capabilities previously thought to require massive scale. While recent work optimizes \textit{how} to scale, we demonstrate that \textit{whether to scale} depends critically on data geometry. ClaD’s innovations -- manifold-aware training, Mahalanobis contrast, and the decision algorithm -- prove that for nuanced language understanding tasks, modeling latent structure trumps brute-force scaling.

Our work does not negate scaling, but expands the efficiency frontier, showing that geometric principles can supplant scale and provide a complementary pathway for real-world applications. 
As AI research increasingly prioritizes \textit{efficiency} alongside performance -- whether through scaling laws, sparsity, or geometric learning -- our findings position the geometric understanding of data as a foundational pillar of sustainable NLP.

\section*{Limitations}
\label{sec:limitations}
While our proposed \textbf{Cla}ss \textbf{D}istillation (ClaD) paradigm demonstrates consistently strong performance and efficiency across sarcasm, metaphor, and sexism detection tasks -- outperforming smaller Transformer models with equal training, and LLMs with in limited resouce settings -- several limitations should be acknowledged. First, although we tested ClaD on diverse tasks encompassing figurative and deviant language, the chosen benchmarks (Sarcasm Headlines, TroFi, and CMSB) may not fully capture the richness of real-world scenarios, and more domain-specific or multilingual tasks could present additional linguistic and cultural nuances not addressed in our current evaluation. Whether our approach can be generalized to specialized domains like legal or clinical tasks also remains to be seen. Second, ClaD relies on the ability of the target class manifold to be modeled as a multivariate normal distribution. While our experiments suggest that training can nudge embeddings closer to a normal manifold, this assumption may not hold universally: certain representations or highly imbalanced corpora may exhibit multimodal or heavy-tailed distributions that deviate substantially from normality, potentially affecting performance. Third, although ClaD’s fast convergence leads to significant computational savings compared to multi-epoch fine-tuning, maintaining a dynamically updated covariance matrix in the Mahalanobis distance computation can be memory-intensive for very large datasets. 
Further advances in this line of research will likely require more memory-efficient approximations or low-rank updates.

These limitations may be addressed by exploring alternative distributional assumptions (\textit{e.g.}, Gaussian mixtures) to accommodate more complex embedding spaces, conducting broader evaluations across languages and task domains, and developing lightweight variants of Mahalanobis-based training to reduce the memory overhead. They have the potential to further enhance ClaD’s versatility and impact in real-world applications.

\section*{Ethics Statement}

This work adheres to ethical standards in NLP research by ensuring transparency, reproducibility, and fairness in our experiments. Our study does not involve human subjects or sensitive data, and all datasets used are publicly available with appropriate licenses. While our findings highlight efficiency gaps in large-scale language models, we acknowledge that their broader societal impacts, including biases and potential misuse, require further investigation. We encourage responsible deployment of our proposed methods and emphasize the need for continued ethical scrutiny in model development and evaluation.

\bibliography{custom, banerjee}

\begin{thebibliography}{64}
\providecommand{\natexlab}[1]{#1}

\bibitem[{Anderson and Darling(1952)}]{Anderson1952AsymptoticTO}
Theodore~W. Anderson and Donald Darling. 1952.
\newblock \href {https://api.semanticscholar.org/CorpusID:120541257} {Asymptotic theory of certain "goodness of fit" criteria based on stochastic processes}.
\newblock \emph{Annals of Mathematical Statistics}, 23:193--212.

\bibitem[{Athanasiadou(2024)}]{athanasiadou2024margins}
Angeliki Athanasiadou. 2024.
\newblock \href {https://doi.org/10.1016/j.lingua.2023.103655} {On the margins of figurative thought and language}.
\newblock \emph{Lingua}, 299:103655.

\bibitem[{Bellman(1957)}]{bellman1965dynamic}
Richard Bellman. 1957.
\newblock \emph{{Dynamic Programming}}.
\newblock Princeton University Press.

\bibitem[{Bender et~al.(2021)Bender, Gebru, McMillan-Major, and Shmitchell}]{bender2021parrots}
Emily~M. Bender, Timnit Gebru, Angelina McMillan-Major, and Shmargaret Shmitchell. 2021.
\newblock \href {https://doi.org/10.1145/3442188.3445922} {{On the Dangers of Stochastic Parrots: Can Language Models Be Too Big?}}
\newblock In \emph{Proceedings of the 2021 ACM Conference on Fairness, Accountability, and Transparency}, FAccT '21, page 610–623, New York, NY, USA. Association for Computing Machinery.

\bibitem[{Birke and Sarkar(2006)}]{birke2006clustering}
Julia Birke and Anoop Sarkar. 2006.
\newblock \href {https://aclanthology.org/E06-1042} {{A Clustering Approach for Nearly Unsupervised Recognition of Nonliteral Language}}.
\newblock In \emph{11th Conference of the {E}uropean Chapter of the Association for Computational Linguistics}, pages 329--336, Trento, Italy. Association for Computational Linguistics.

\bibitem[{Bose and Su(2022)}]{bose2022deep}
Saugata Bose and Guoxin Su. 2022.
\newblock Deep one-class hate speech detection model.
\newblock In \emph{Proceedings of the Thirteenth Language Resources and Evaluation Conference}, pages 7040--7048.

\bibitem[{Cer et~al.(2017)Cer, Diab, Agirre, Lopez-Gazpio, and Specia}]{cer2017semeval}
Daniel Cer, Mona Diab, Eneko Agirre, I{\~n}igo Lopez-Gazpio, and Lucia Specia. 2017.
\newblock \href {https://doi.org/10.18653/v1/S17-2001} {{SemEval-2017 Task 1: Semantic Textual Similarity Multilingual and Crosslingual Focused Evaluation}}.
\newblock In \emph{Proceedings of the 11th International Workshop on Semantic Evaluation ({S}em{E}val-2017)}, pages 1--14. Association for Computational Linguistics.

\bibitem[{Chakrabarty et~al.(2022)Chakrabarty, Saakyan, Ghosh, and Muresan}]{chakrabarty2022flute}
Tuhin Chakrabarty, Arkadiy Saakyan, Debanjan Ghosh, and Smaranda Muresan. 2022.
\newblock {FLUTE: Figurative Language Understanding through Textual Explanations}.
\newblock \emph{arXiv}, 2205.12404v3.

\bibitem[{Chalapathy and Chawla(2019)}]{chalapathy2019deep}
Raghavendra Chalapathy and Sanjay Chawla. 2019.
\newblock \href {https://doi.org/10.48550/arXiv.1901.03407} {{Deep Learning for Anomaly Detection: A Survey}}.
\newblock \emph{arXiv}, 1901.03407.

\bibitem[{Devlin et~al.(2019)Devlin, Chang, Lee, and Toutanova}]{devlin2019bert}
Jacob Devlin, Ming-Wei Chang, Kenton Lee, and Kristina Toutanova. 2019.
\newblock \href {https://doi.org/10.18653/v1/N19-1423} {{BERT}: Pre-training of deep bidirectional transformers for language understanding}.
\newblock In \emph{Proceedings of the 2019 Conference of the North {A}merican Chapter of the Association for Computational Linguistics: Human Language Technologies, Volume 1 (Long and Short Papers)}, pages 4171--4186, Minneapolis, Minnesota. Association for Computational Linguistics.

\bibitem[{Ethayarajh(2019)}]{ethayarajh2019contextual}
Kawin Ethayarajh. 2019.
\newblock \href {https://doi.org/10.18653/v1/D19-1006} {How contextual are contextualized word representations? {C}omparing the geometry of {BERT}, {ELM}o, and {GPT}-2 embeddings}.
\newblock In \emph{Proceedings of the 2019 Conference on Empirical Methods in Natural Language Processing and the 9th International Joint Conference on Natural Language Processing (EMNLP-IJCNLP)}, pages 55--65, Hong Kong, China. Association for Computational Linguistics.

\bibitem[{Fortuna and Nunes(2018)}]{fortuna2018survey}
Paula Fortuna and S{\'e}rgio Nunes. 2018.
\newblock {A Survey on Automatic Detection of Hate Speech in Text}.
\newblock \emph{ACM Computing Surveys}, 51(4):1--30.

\bibitem[{Gao et~al.(2021)Gao, Yao, and Chen}]{gao2021simcse}
Tianyu Gao, Xingcheng Yao, and Danqi Chen. 2021.
\newblock \href {https://doi.org/10.18653/V1/2021.EMNLP-MAIN.552} {{SimCSE: Simple Contrastive Learning of Sentence Embeddings}}.
\newblock In \emph{Proceedings of the 2021 Conference on Empirical Methods in Natural Language Processing}, pages 6894--6910. Association for Computational Linguistics.

\bibitem[{Ge et~al.(2023)Ge, Mao, and Cambria}]{ge2023survey}
Mengshi Ge, Rui Mao, and Erik Cambria. 2023.
\newblock \href {https://doi.org/10.1007/s10462-023-10564-7} {{A survey on computational metaphor processing techniques: from identification, interpretation, generation to application}}.
\newblock \emph{Artif. Intell. Rev.}, 56(2):1829--1895.

\bibitem[{Ghosh et~al.(2020)Ghosh, Vajpayee, and Muresan}]{ghosh2020report}
Debanjan Ghosh, Avijit Vajpayee, and Smaranda Muresan. 2020.
\newblock {A Report on the 2020 Sarcasm Detection Shared Task}.

\bibitem[{Gregory et~al.(2020)Gregory, Li, Mohammadi, Tarn, Draelos, and Rudin}]{gregory2020transformer}
Hunter Gregory, Steven Li, Pouya Mohammadi, Natalie Tarn, Rachel Draelos, and Cynthia Rudin. 2020.
\newblock A transformer approach to contextual sarcasm detection in twitter.
\newblock In \emph{Proceedings of the second workshop on figurative language processing}, pages 270--275.

\bibitem[{Guest et~al.(2021)Guest, Vidgen, Mittos, Sastry, Tyson, and Margetts}]{guest2021expert}
Ella Guest, Bertie Vidgen, Alexandros Mittos, Nishanth Sastry, Gareth Tyson, and Helen Margetts. 2021.
\newblock \href {https://doi.org/10.18653/v1/2021.eacl-main.114} {{An Expert Annotated Dataset for the Detection of Online Misogyny}}.
\newblock In \emph{Proceedings of the 16th Conference of the European Chapter of the Association for Computational Linguistics: Main Volume}, pages 1336--1350, Online. Association for Computational Linguistics.

\bibitem[{Guo et~al.(2025)Guo, Yang, Zhang, Song, Zhang, Xu et~al.}]{deepseek_r1}
Daya Guo, Dejian Yang, Haowei Zhang, Junxiao Song, Ruoyu Zhang, Runxin Xu, et~al. 2025.
\newblock \href {https://doi.org/10.48550/arXiv.2501.12948} {{DeepSeek-R1: Incentivizing Reasoning Capability in LLMs via Reinforcement Learning}}.
\newblock \emph{arXiv}, 2501.12948.

\bibitem[{Gururangan et~al.(2018)Gururangan, Swayamdipta, Levy, Schwartz, Bowman, and Smith}]{gururangan2018annotation}
Suchin Gururangan, Swabha Swayamdipta, Omer Levy, Roy Schwartz, Samuel Bowman, and Noah~A. Smith. 2018.
\newblock \href {https://doi.org/10.18653/v1/N18-2017} {Annotation artifacts in natural language inference data}.
\newblock In \emph{Proceedings of the 2018 Conference of the North {A}merican Chapter of the Association for Computational Linguistics: Human Language Technologies, Volume 2 (Short Papers)}, pages 107--112, New Orleans, Louisiana. Association for Computational Linguistics.

\bibitem[{He et~al.(2021)He, Liu, Gao, and Chen}]{he2021deberta}
Pengcheng He, Xiaodong Liu, Jianfeng Gao, and Weizhu Chen. 2021.
\newblock \href {https://openreview.net/forum?id=XPZIaotutsD} {{DeBERTa: Decoding-Enhanced BERT With Disentangled Attention}}.
\newblock In \emph{International Conference on Learning Representations}.

\bibitem[{Henze and Zirkler(1990)}]{Henze1990ACO}
Norbert Henze and Bernd Zirkler. 1990.
\newblock \href {https://api.semanticscholar.org/CorpusID:120328121} {A class of invariant consistent tests for multivariate normality}.
\newblock \emph{Communications in Statistics-theory and Methods}, 19:3595--3617.

\bibitem[{Hotelling(1936)}]{hotelling1933pca}
Harold Hotelling. 1936.
\newblock \href {https://doi.org/10.2307/2333955} {{Relations between two sets of variates}}.
\newblock \emph{Biometrika}, 28(3/4):321 -- 377.

\bibitem[{Hu et~al.(2022)Hu, Shen, Wallis, Allen-Zhu, Li, Wang, Wang, and Chen}]{hu2022lora}
Edward~J. Hu, Yelong Shen, Phillip Wallis, Zeyuan Allen-Zhu, Yuanzhi Li, Shean Wang, Lu~Wang, and Weizhu Chen. 2022.
\newblock {LoRA: Low-Rank Adaptation of Large Language Models}.
\newblock In \emph{International Conference on Learning Representations}.

\bibitem[{Javaheripi et~al.(2023)Javaheripi, Bubeck, Abdin, Aneja, Bubeck, Mendes, Chen, Del~Giorno, Eldan, Gopi et~al.}]{javaheripi2023phi}
Mojan Javaheripi, S{\'e}bastien Bubeck, Marah Abdin, Jyoti Aneja, Sebastien Bubeck, Caio C{\'e}sar~Teodoro Mendes, Weizhu Chen, Allie Del~Giorno, Ronen Eldan, Sivakanth Gopi, et~al. 2023.
\newblock {Phi-2: The surprising power of small language models}.
\newblock \emph{Microsoft Research Blog}.

\bibitem[{Kasparian(2013)}]{kasparian2013hemispheric}
Kristina Kasparian. 2013.
\newblock \href {https://doi.org/10.1016/j.jneuroling.2012.07.001} {{H}emispheric differences in figurative language processing: {C}ontributions of neuroimaging methods and challenges in reconciling current empirical findings}.
\newblock \emph{Journal of Neurolinguistics}, 26(1):1--21.

\bibitem[{Kiska(2012)}]{kiska2012hate}
Roger Kiska. 2012.
\newblock {Hate Speech: A Comparison between the European Court of Human Rights and the United States Supreme Court Jurisprudence }.
\newblock \emph{Regent Univerity Law Review}, 25:107.

\bibitem[{Kumar and Anand(2020)}]{kumar2020transformers}
Amardeep Kumar and Vivek Anand. 2020.
\newblock Transformers on sarcasm detection with context.
\newblock In \emph{Proceedings of the second workshop on figurative language processing}, pages 88--92.

\bibitem[{Lai and Nissim(2024)}]{huiyuan2024survey}
Huiyuan Lai and Malvina Nissim. 2024.
\newblock \href {https://doi.org/10.1145/3654795} {{A Survey on Automatic Generation of Figurative Language: From Rule-based Systems to Large Language Models}}.
\newblock \emph{ACM Comput. Surv.}, 56(10).

\bibitem[{Lan et~al.(2020)Lan, Chen, Goodman, Gimpel, Sharma, and Soricut}]{lan2020albert}
Zhenzhong Lan, Mingda Chen, Sebastian Goodman, Kevin Gimpel, Piyush Sharma, and Radu Soricut. 2020.
\newblock \href {https://iclr.cc/virtual_2020/poster_H1eA7AEtvS.html} {{ALBERT: A Lite BERT for Self-supervised Learning of Language Representations}}.
\newblock In \emph{International Conference on Learning Representations (ICLR)}.

\bibitem[{Lemmens et~al.(2020)Lemmens, Burtenshaw, Lotfi, Markov, and Daelemans}]{lemmens2020sarcasm}
Jens Lemmens, Ben Burtenshaw, Ehsan Lotfi, Ilia Markov, and Walter Daelemans. 2020.
\newblock Sarcasm detection using an ensemble approach.
\newblock In \emph{proceedings of the second workshop on figurative language processing}, pages 264--269.

\bibitem[{Liu et~al.(2024)Liu, Feng, Xue, Wang, Wu, Lu, Zhao et~al.}]{deepseek_v3}
Aixin Liu, Bei Feng, Bing Xue, Bingxuan Wang, Bochao Wu, Chengda Lu, Chenggang Zhao, et~al. 2024.
\newblock {DeepSeek-V3 Technical Report}.
\newblock Technical report, DeepSeek-AI.

\bibitem[{Liu et~al.(2008)Liu, Ting, and Zhou}]{liu2008isolation}
Fei~Tony Liu, Kai~Ming Ting, and Zhi-Hua Zhou. 2008.
\newblock \href {https://doi.org/10.1109/ICDM.2008.17} {{Isolation Forest}}.
\newblock In \emph{2008 Eighth IEEE International Conference on Data Mining}, pages 413--422.

\bibitem[{Liu et~al.(2010)Liu, Ting, and Zhou}]{liu2010detecting}
Fei~Tony Liu, Kai~Ming Ting, and Zhi-Hua Zhou. 2010.
\newblock \href {https://doi.org/10.1007/978-3-642-15883-4_18} {{On Detecting Clustered Anomalies Using SCiForest}}.
\newblock In \emph{Machine Learning and Knowledge Discovery in Databases}, pages 274--290, Berlin, Heidelberg. Springer Berlin Heidelberg.

\bibitem[{Mahalanobis(1936)}]{mahalanobis1936}
Prasanta~Chandra Mahalanobis. 1936.
\newblock {On the Generalized Distance in Statistics}.
\newblock \emph{Proceedings of the National Institute of Sciences of India}, 2(1):49--55.

\bibitem[{Misra and Arora(2019)}]{misra2019sarcasm}
Rishabh Misra and Prahal Arora. 2019.
\newblock \href {https://doi.org/10.48550/arXiv.1908.07414} {{Sarcasm Detection using Hybrid Neural Network}}.
\newblock \emph{arXiv}, 1908.07414.

\bibitem[{Misra and Arora(2023)}]{misra2023Sarcasm}
Rishabh Misra and Prahal Arora. 2023.
\newblock \href {https://doi.org/10.1016/j.aiopen.2023.01.001} {Sarcasm detection using news headlines dataset}.
\newblock \emph{AI Open}, 4:13--18.

\bibitem[{Noumir et~al.(2012)Noumir, Honeine, and Richard}]{noumir2012occ}
Zineb Noumir, Paul Honeine, and C\'edue Richard. 2012.
\newblock \href {https://doi.org/10.1109/ISIT.2012.6283685} {{On simple one-class classification methods}}.
\newblock In \emph{2012 IEEE International Symposium on Information Theory Proceedings}, pages 2022--2026.

\bibitem[{Oraby et~al.(2016)Oraby, Harrison, Reed, Hernandez, Riloff, and Walker}]{oraby2016creating}
Shereen Oraby, Vrindavan Harrison, Lena Reed, Ernesto Hernandez, Ellen Riloff, and Marilyn Walker. 2016.
\newblock \href {https://doi.org/10.18653/v1/W16-3604} {{Creating and Characterizing a Diverse Corpus of Sarcasm in Dialogue}}.
\newblock In \emph{Proceedings of the 17th Annual Meeting of the Special Interest Group on Discourse and Dialogue}, pages 31--41, Los Angeles. Association for Computational Linguistics.

\bibitem[{Pearson(1901)}]{pearson1901pca}
Karl Pearson. 1901.
\newblock \href {https://doi.org/10.1080/14786440109462720} {{On Lines and Planes of Closest Fit to Systems of Points in Space}}.
\newblock \emph{Philosophical Magazine}, 2(11):559 -- 572.

\bibitem[{Radford et~al.(2019)Radford, Wu, Child, Luan, Amodei, Sutskever et~al.}]{radford2019language}
Alec Radford, Jeffrey Wu, Rewon Child, David Luan, Dario Amodei, Ilya Sutskever, et~al. 2019.
\newblock {Language Models are Unsupervised Multitask Learners}.
\newblock \emph{OpenAI blog}, 1(8):9.

\bibitem[{Reimers and Gurevych(2019)}]{reimers2019sentence}
Nils Reimers and Iryna Gurevych. 2019.
\newblock \href {https://doi.org/10.18653/V1/D19-1410} {{Sentence-BERT: Sentence Embeddings using Siamese BERT-Networks}}.
\newblock In \emph{Proceedings of the 2019 Conference on Empirical Methods in Natural Language Processing and the 9th International Joint Conference on Natural Language Processing}, pages 3980--3990. Association for Computational Linguistics.

\bibitem[{Riloff et~al.(2013)Riloff, Qadir, Surve, De~Silva, Gilbert, and Huang}]{riloff2013sarcasm}
Ellen Riloff, Ashequl Qadir, Prafulla Surve, Lalindra De~Silva, Nathan Gilbert, and Ruihong Huang. 2013.
\newblock \href {https://aclanthology.org/D13-1066} {{Sarcasm as Contrast between a Positive Sentiment and Negative Situation}}.
\newblock In \emph{Proceedings of the 2013 Conference on Empirical Methods in Natural Language Processing}, pages 704--714, Seattle, Washington, USA. Association for Computational Linguistics.

\bibitem[{Rogers et~al.(2021)Rogers, Kovaleva, and Rumshisky}]{rogers2021primer}
Anna Rogers, Olga Kovaleva, and Anna Rumshisky. 2021.
\newblock \href {https://doi.org/10.1162/tacl_a_00349} {{A Primer in BERTology: What We Know About How BERT Works}}.
\newblock \emph{Transactions of the Association for Computational Linguistics}, 8:842--866.

\bibitem[{Rosenfeld(2002)}]{rosenfeld2002hate}
Michel Rosenfeld. 2002.
\newblock {Hate Speech in Constitutional Jurisprudence: A Comparative Analysis}.
\newblock \emph{Cardozo Law Review}, 24:1523.

\bibitem[{Ruff et~al.(2018)Ruff, G{\"{o}}rnitz, Deecke, Siddiqui, Vandermeulen, Binder, M{\"{u}}ller, and Kloft}]{DBLP:conf/icml/RuffGDSVBMK18}
Lukas Ruff, Nico G{\"{o}}rnitz, Lucas Deecke, Shoaib~Ahmed Siddiqui, Robert~A. Vandermeulen, Alexander Binder, Emmanuel M{\"{u}}ller, and Marius Kloft. 2018.
\newblock \href {http://proceedings.mlr.press/v80/ruff18a.html} {Deep one-class classification}.
\newblock In \emph{Proceedings of the 35th International Conference on Machine Learning, {ICML} 2018, Stockholmsm{\"{a}}ssan, Stockholm, Sweden, July 10-15, 2018}, volume~80 of \emph{Proceedings of Machine Learning Research}, pages 4390--4399. {PMLR}.

\bibitem[{Saakyan et~al.(2022)Saakyan, Chakrabarty, Ghosh, and Muresan}]{saakyan2022report}
Arkadiy Saakyan, Tuhin Chakrabarty, Debanjan Ghosh, and Smaranda Muresan. 2022.
\newblock \href {https://doi.org/10.18653/v1/2022.flp-1.26} {{A Report on the FigLang 2022 Shared Task on Understanding Figurative Language}}.
\newblock In \emph{Proceedings of the 3rd Workshop on Figurative Language Processing (FLP)}, pages 178--183, Abu Dhabi, United Arab Emirates (Hybrid). Association for Computational Linguistics.

\bibitem[{Sabokrou et~al.(2018)Sabokrou, Khalooei, Fathy, and Adeli}]{DBLP:conf/cvpr/SabokrouKFA18}
Mohammad Sabokrou, Mohammad Khalooei, Mahmood Fathy, and Ehsan Adeli. 2018.
\newblock \href {https://doi.org/10.1109/CVPR.2018.00356} {Adversarially learned one-class classifier for novelty detection}.
\newblock In \emph{2018 {IEEE} Conference on Computer Vision and Pattern Recognition, {CVPR} 2018, Salt Lake City, UT, USA, June 18-22, 2018}, pages 3379--3388. Computer Vision Foundation / {IEEE} Computer Society.

\bibitem[{Safi~Samghabadi et~al.(2020)Safi~Samghabadi, Patwa, PYKL, Mukherjee, Das, and Solorio}]{samghabadi2020aggression}
Niloofar Safi~Samghabadi, Parth Patwa, Srinivas PYKL, Prerana Mukherjee, Amitava Das, and Thamar Solorio. 2020.
\newblock \href {https://aclanthology.org/2020.trac-1.20} {{Aggression and Misogyny Detection using BERT: A Multi-Task Approach}}.
\newblock In \emph{Proceedings of the Second Workshop on Trolling, Aggression and Cyberbullying}, pages 126--131, Marseille, France. European Language Resources Association (ELRA).

\bibitem[{Samory et~al.(2021)Samory, Sen, Kohne, Flöck, and Wagner}]{samory2021sexism}
Mattia Samory, Indira Sen, Julian Kohne, Fabian Flöck, and Claudia Wagner. 2021.
\newblock \href {https://doi.org/10.1609/icwsm.v15i1.18085} {{“Call me sexist, but...” : Revisiting Sexism Detection Using Psychological Scales and Adversarial Samples}}.
\newblock \emph{Proceedings of the International AAAI Conference on Web and Social Media}, 15(1):573--584.

\bibitem[{Sanh et~al.(2019)Sanh, Debut, Chaumond, and Wolf}]{sanh2019distilbert}
Victor Sanh, Lysandre Debut, Julien Chaumond, and Thomas Wolf. 2019.
\newblock \href {https://www.emc2-ai.org/assets/docs/neurips-19/emc2-neurips19-paper-33.pdf} {{DistilBERT, a distilled version of BERT: smaller, faster, cheaper and lighter}}.
\newblock In \emph{5th Workshop on Energy Efficient Machine Learning and Cognitive Computing at NeurIPS 2019}.

\bibitem[{Sch{\"{o}}lkopf et~al.(2001)Sch{\"{o}}lkopf, Platt, Shawe{-}Taylor, Smola, and Williamson}]{DBLP:journals/neco/ScholkopfPSSW01}
Bernhard Sch{\"{o}}lkopf, John~C. Platt, John Shawe{-}Taylor, Alexander~J. Smola, and Robert~C. Williamson. 2001.
\newblock \href {https://doi.org/10.1162/089976601750264965} {Estimating the support of a high-dimensional distribution}.
\newblock \emph{Neural Comput.}, 13(7):1443--1471.

\bibitem[{Sun et~al.(2023)Sun, Li, Li, Wu, Guo, Zhang, and Wang}]{DBLP:conf/emnlp/SunL0WGZ023}
Xiaofei Sun, Xiaoya Li, Jiwei Li, Fei Wu, Shangwei Guo, Tianwei Zhang, and Guoyin Wang. 2023.
\newblock \href {https://doi.org/10.18653/V1/2023.FINDINGS-EMNLP.603} {Text classification via large language models}.
\newblock In \emph{Findings of the Association for Computational Linguistics: {EMNLP} 2023, Singapore, December 6-10, 2023}, pages 8990--9005. Association for Computational Linguistics.

\bibitem[{Timkey and van Schijndel(2021)}]{timkey2021bark}
William Timkey and Marten van Schijndel. 2021.
\newblock \href {https://doi.org/10.18653/v1/2021.emnlp-main.372} {{All Bark and No Bite: Rogue Dimensions in Transformer Language Models Obscure Representational Quality}}.
\newblock In \emph{Proceedings of the 2021 Conference on Empirical Methods in Natural Language Processing}, pages 4527--4546, Online and Punta Cana, Dominican Republic. Association for Computational Linguistics.

\bibitem[{van~der Maaten and Hinton(2008)}]{maaten2008tsne}
Laurens van~der Maaten and Geoffrey Hinton. 2008.
\newblock {Visualizing Data Using t-SNE}.
\newblock \emph{Journal of Machine Learning Research}, 9:2579 -- 2605.

\bibitem[{Ververidis and Kotropoulos(2008)}]{ververidis2008gaussian}
Dimitrios Ververidis and Constantine Kotropoulos. 2008.
\newblock Gaussian mixture modeling by exploiting the mahalanobis distance.
\newblock \emph{IEEE transactions on signal processing}, 56(7):2797--2811.

\bibitem[{Wachowiak and Gromann(2023)}]{wachowiak2023gpt}
Lennart Wachowiak and Dagmar Gromann. 2023.
\newblock \href {https://doi.org/10.18653/v1/2023.acl-long.58} {Does {GPT}-3 grasp metaphors? identifying metaphor mappings with generative language models}.
\newblock In \emph{Proceedings of the 61st Annual Meeting of the Association for Computational Linguistics (Volume 1: Long Papers)}, pages 1018--1032, Toronto, Canada. Association for Computational Linguistics.

\bibitem[{Wan et~al.(2021)Wan, Lin, Du, Shen, and Zhang}]{wan2021enhancing}
Hai Wan, Jinxia Lin, Jianfeng Du, Dawei Shen, and Manrong Zhang. 2021.
\newblock \href {https://doi.org/10.18653/v1/2021.findings-acl.173} {{Enhancing Metaphor Detection by Gloss-based Interpretations}}.
\newblock In \emph{Findings of the Association for Computational Linguistics: ACL-IJCNLP 2021}, pages 1971--1981, Online. Association for Computational Linguistics.

\bibitem[{Wan et~al.(2020)Wan, Ahrens, Chersoni, Jiang, Su, Xiang, and Huang}]{wan2020using}
Mingyu Wan, Kathleen Ahrens, Emmanuele Chersoni, Menghan Jiang, Qi~Su, Rong Xiang, and Chu-Ren Huang. 2020.
\newblock Using conceptual norms for metaphor detection.
\newblock In \emph{Proceedings of the Second Workshop on Figurative Language Processing}, pages 104--109.

\bibitem[{Wang et~al.(2023)Wang, Pang, and Lin}]{DBLP:journals/corr/abs-2312-01044}
Zhiqiang Wang, Yiran Pang, and Yanbin Lin. 2023.
\newblock \href {https://doi.org/10.48550/ARXIV.2312.01044} {Large language models are zero-shot text classifiers}.
\newblock \emph{CoRR}, abs/2312.01044.

\bibitem[{Wilks(1962)}]{wilks1962mathematical}
Samuel~S Wilks. 1962.
\newblock Mathematical statistics. a wiley publication in mathematical statistics john wiley \& sons.
\newblock \emph{Inc., New York-London}, pages 0173--45805.

\bibitem[{Yang et~al.(2019)Yang, Dai, Yang, Carbonell, Salakhutdinov, and Le}]{yang2019xlnet}
Zhilin Yang, Zihang Dai, Yiming Yang, Jaime Carbonell, Russ~R Salakhutdinov, and Quoc~V Le. 2019.
\newblock \href {https://proceedings.neurips.cc/paper_files/paper/2019/file/dc6a7e655d7e5840e66733e9ee67cc69-Paper.pdf} {{XLNet: Generalized Autoregressive Pretraining for Language Understanding}}.
\newblock In \emph{Advances in Neural Information Processing Systems}, volume~32. Curran Associates, Inc.

\bibitem[{Yin and Zubiaga(2021)}]{yin2021towards}
Wenjie Yin and Arkaitz Zubiaga. 2021.
\newblock Towards generalisable hate speech detection: a review on obstacles and solutions.
\newblock \emph{PeerJ Computer Science}, 7:e598.

\bibitem[{Zhang et~al.(2022)Zhang, Roller, Goyal, Artetxe, Chen, Chen, Dewan, Diab, Li, Lin, Mihaylov, Ott, Shleifer, Shuster, Simig, Koura, Sridhar, Wang, and Zettlemoyer}]{zhang2022opt}
Susan Zhang, Stephen Roller, Naman Goyal, Mikel Artetxe, Moya Chen, Shuohui Chen, Christopher Dewan, Mona Diab, Xian Li, Xi~Victoria Lin, Todor Mihaylov, Myle Ott, Sam Shleifer, Kurt Shuster, Daniel Simig, Punit~Singh Koura, Anjali Sridhar, Tianlu Wang, and Luke Zettlemoyer. 2022.
\newblock {OPT: Open Pre-trained Transformer Language Models}.
\newblock \emph{arXiv}, 2205.01068.

\bibitem[{Zhou et~al.(2021)Zhou, Yong, Fan, Ren, Song, Diao, Yang, and Lin}]{zhou2021hate}
Xianbing Zhou, Yang Yong, Xiaochao Fan, Ge~Ren, Yunfeng Song, Yufeng Diao, Liang Yang, and Hongfei Lin. 2021.
\newblock Hate speech detection based on sentiment knowledge sharing.
\newblock In \emph{Proceedings of the 59th Annual Meeting of the Association for Computational Linguistics and the 11th International Joint Conference on Natural Language Processing (Volume 1: Long Papers)}, pages 7158--7166.

\end{thebibliography}

\clearpage

\appendix

\begin{figure}[!t]
\centering
\begin{subfigure}[t]{0.46\linewidth}
    \centering
    \includegraphics[width=\linewidth]{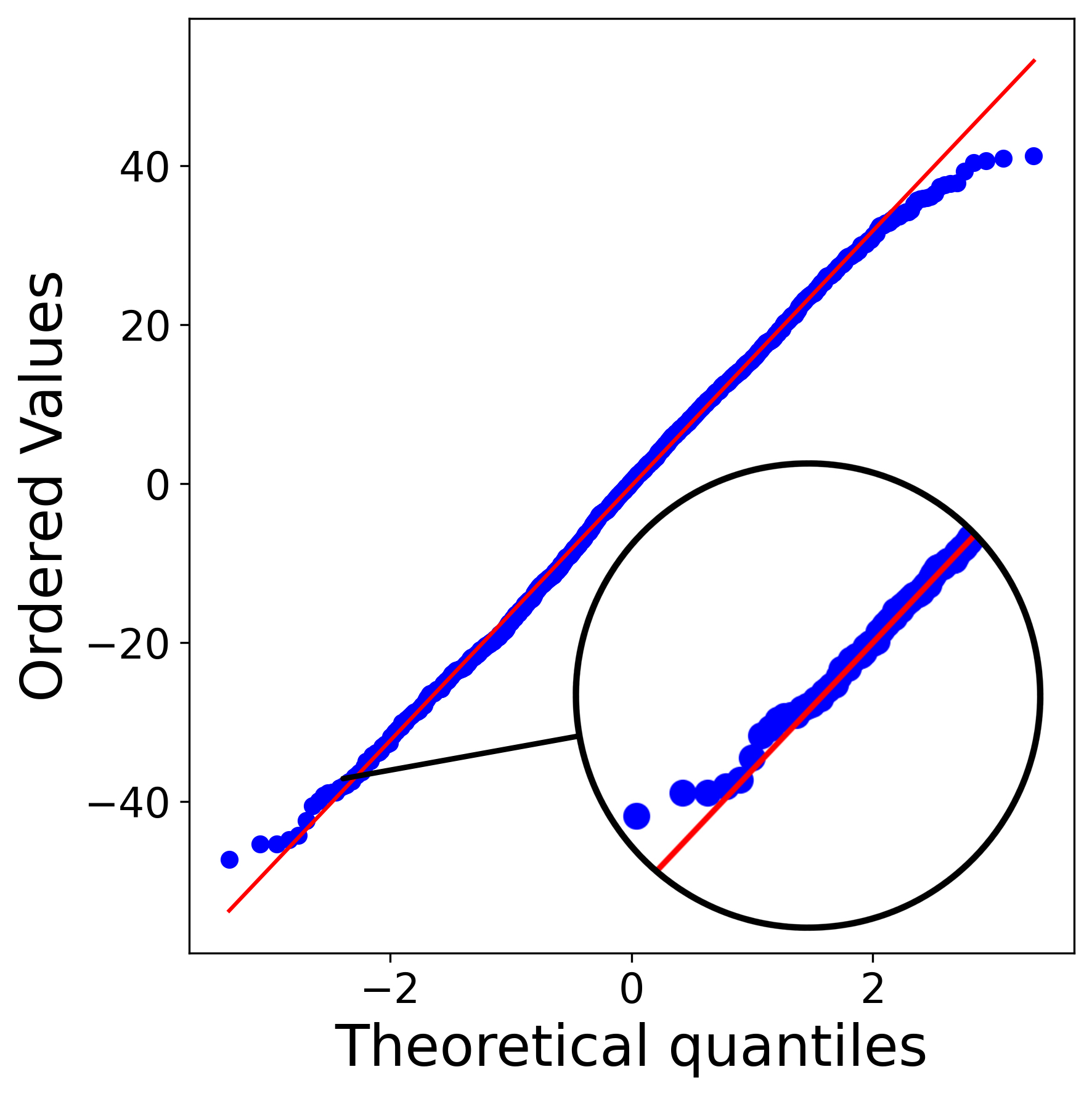}
    \caption{BERT embeddings for the target class, sexism.}
    \label{fig:qqplot5}
\end{subfigure}%
\hfill%
\begin{subfigure}[t]{0.46\linewidth}
    \centering
    \includegraphics[width=\linewidth]{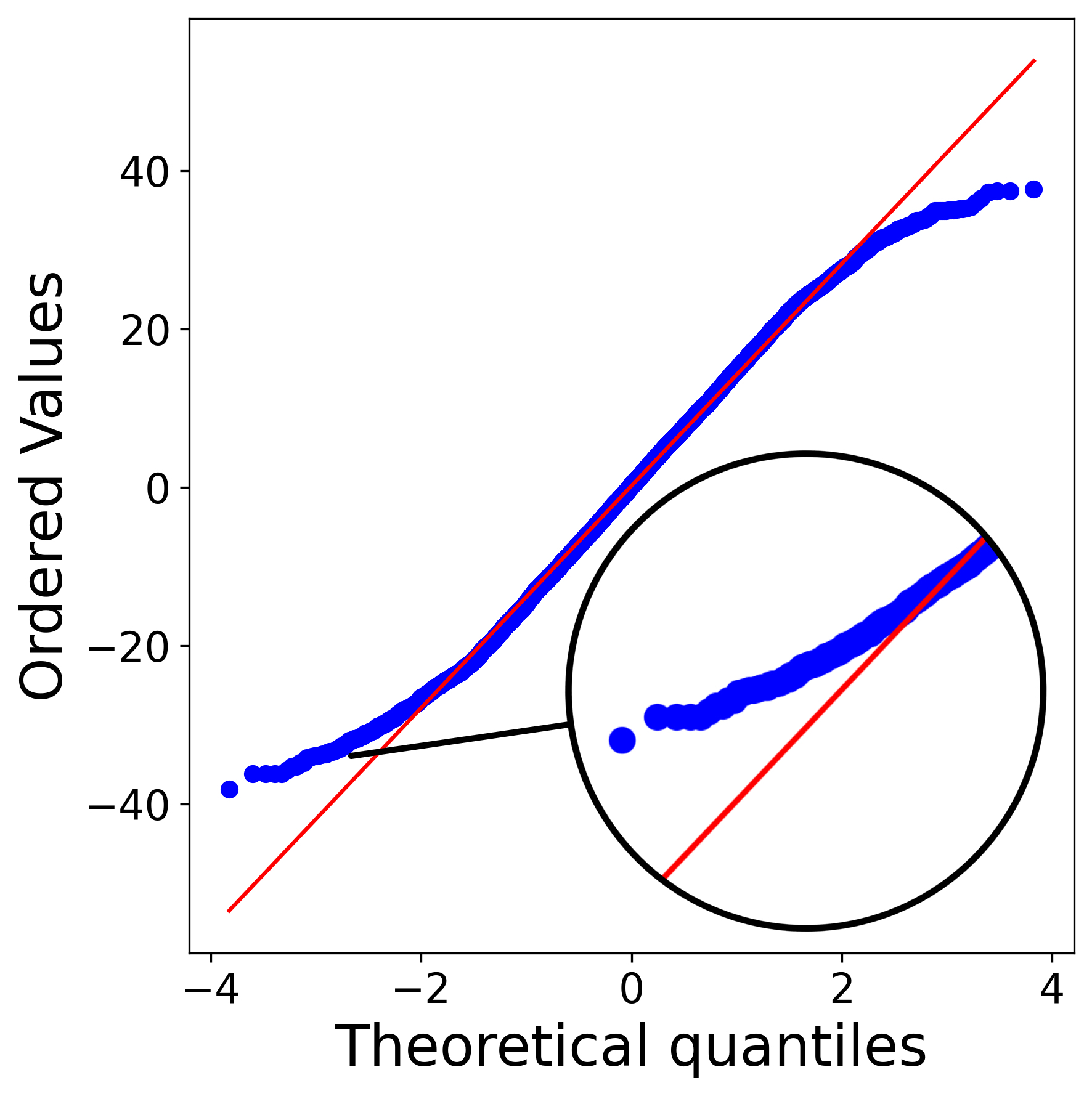}
    \caption{BERT embeddings for the negative class.}
    \label{fig:qqplot4}
\end{subfigure}

\vspace{4pt} 

\begin{subfigure}[t]{0.46\linewidth}
    \centering
    \includegraphics[width=\linewidth]{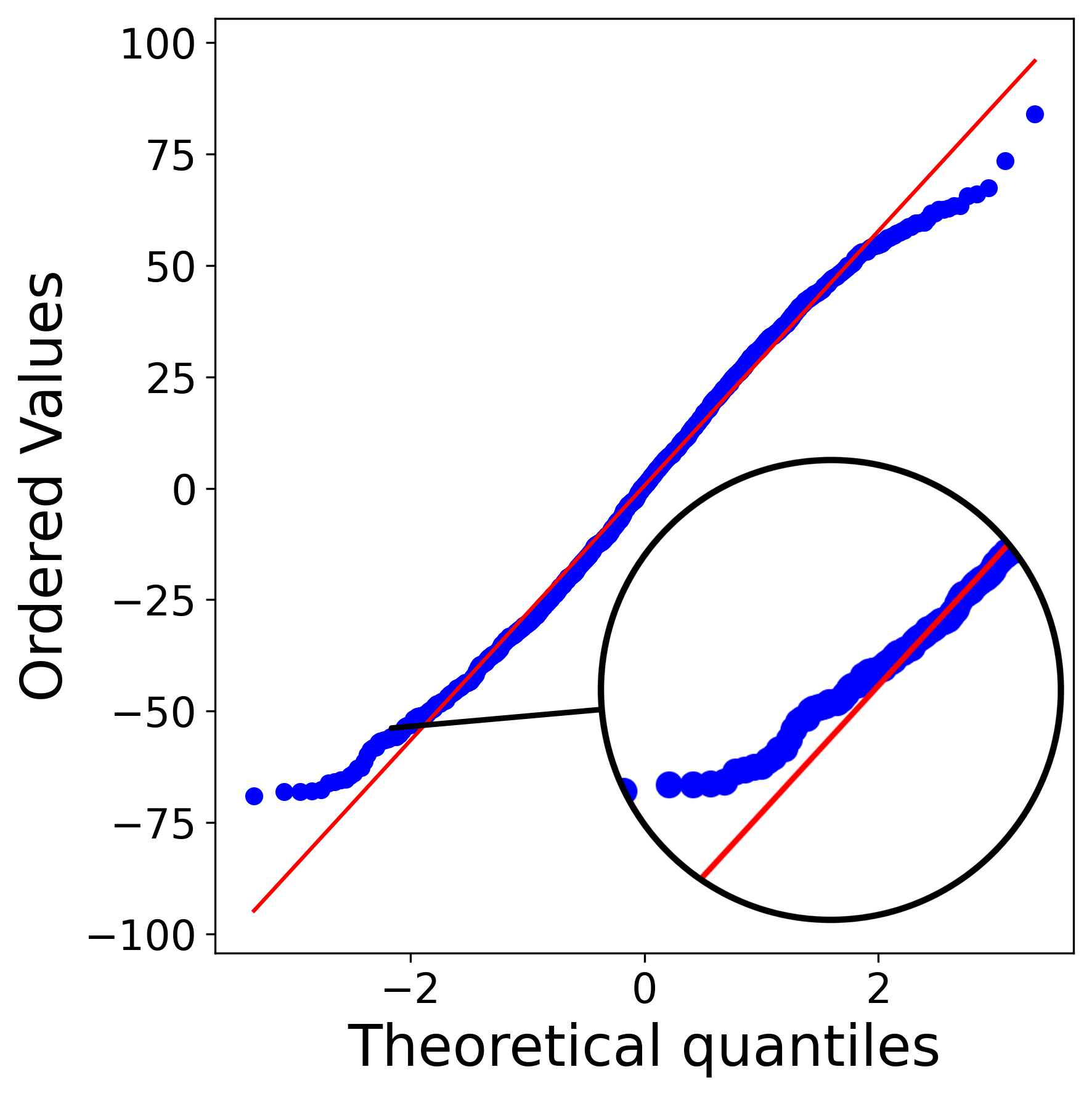}
    \caption{SimCSE embeddings for the target class, sexism.}
    \label{fig:qqplot3}
\end{subfigure}%
\hfill%
\begin{subfigure}[t]{0.46\linewidth}
    \centering
    \includegraphics[width=\linewidth]{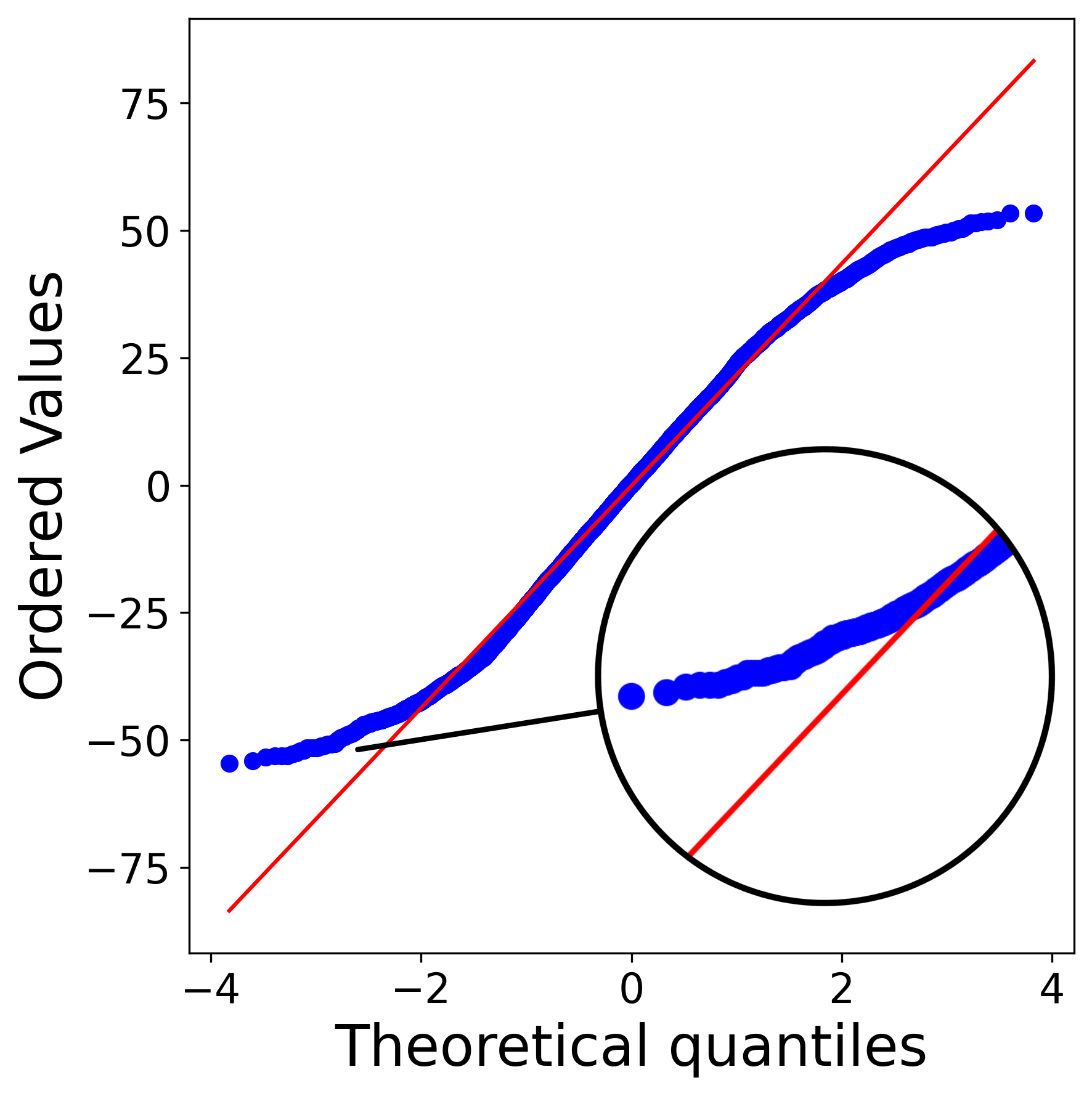}
    \caption{SimCSE embeddings for the negative class.}
    \label{fig:qqplot2}
\end{subfigure}
\caption{Q-Q (quantile-quantile) plots to assess the goodness-of-fit of target and non-target classes in the CMSB corpus for sexism detection. Shown here are the first t-SNE dimensions of pretrained BERT-base (\textit{a} and \textit{b}) and SimCSE (\textit{c} and \textit{d}) embeddings.}
\label{fig:cmsb_qqplots}
\vspace{-11pt}
\end{figure}
\section{A Visual Approach to Goodness-of-Fit in Target Class Distributions}
\label{app:goodness-of-fit}
Our aim is to unveil the distributional properties shared across diverse datasets and tasks, where identifying the minority target class amidst a spectrum of heterogeneous linguistic expressions (with no common pragmatic language function) is paramount. It is well known, however, that there are challenges to such analyses regardless of the manifold structure. As the number of dimensions increase, the volume of space grows exponentially and tests based on density estimation or empirical distance measures can struggle to maintain accuracy due to the increased sparsity and spread of data points.\footnote{The ``curse of dimensionality'' strikes again, as \citet{bellman1965dynamic} presciently described this exponential increase in problem complexity with growing number of dimensions.} Statistical tests also suffer from reduced power in higher dimensions. Moreover, estimating the covariance matrix becomes problematic in higher dimensions (for instance, due to ill-conditioned or singular matrices).

To mitigate these problems, we reduce the number of dimensions using t-SNE~\cite{maaten2008tsne}.\footnote{We also experiment with dimensionality reduction by means of studying anisotropy~\cite{ethayarajh2019contextual} and dominant dimensions as defined by \citet{timkey2021bark}, as well as with principal component analysis~\cite{pearson1901pca, hotelling1933pca}. In each case, the results of the statistical tests of manifold structure are nearly identical. So, for conciseness, we omit the details of these other approaches.} The Quantile-Quantile (Q-Q) plots of the first latent dimension are shown in~\autoref{fig:cmsb_qqplots}, for visual assessment of adherence to a normal distribution. For the sake of brevity, we present only the Q-Q plots for sexism detection using two language models, BERT~\cite{devlin2019bert} and SimCSE~\cite{gao2021simcse}, as similar patterns were observed for the other tasks. 


\section{Configuration}
\label{app:config}

\paragraph{Baseline Models:}
Four baseline models were fine-tuned -- SimCSE, ALBERT, XLNet, and DistilBERT -- on three datasets using a consistent set of configurations. The models were trained with a per-device batch size of 16 for both training (up to 5 epochs) and evaluation (at the end of each epoch). The learning rate is set to $1 \times 10^{-5}$ for all models, with 50 warm-up steps and a weight decay of 0.01. For tokenization, inputs were padded and truncated to a maximum sequence length of 512. We used the Adam optimizer for parameter updates.

\begin{figure*}[!t]
\centering
\begin{subfigure}[t]{0.33\textwidth}
  \centering
  \includegraphics[width=0.98\textwidth]{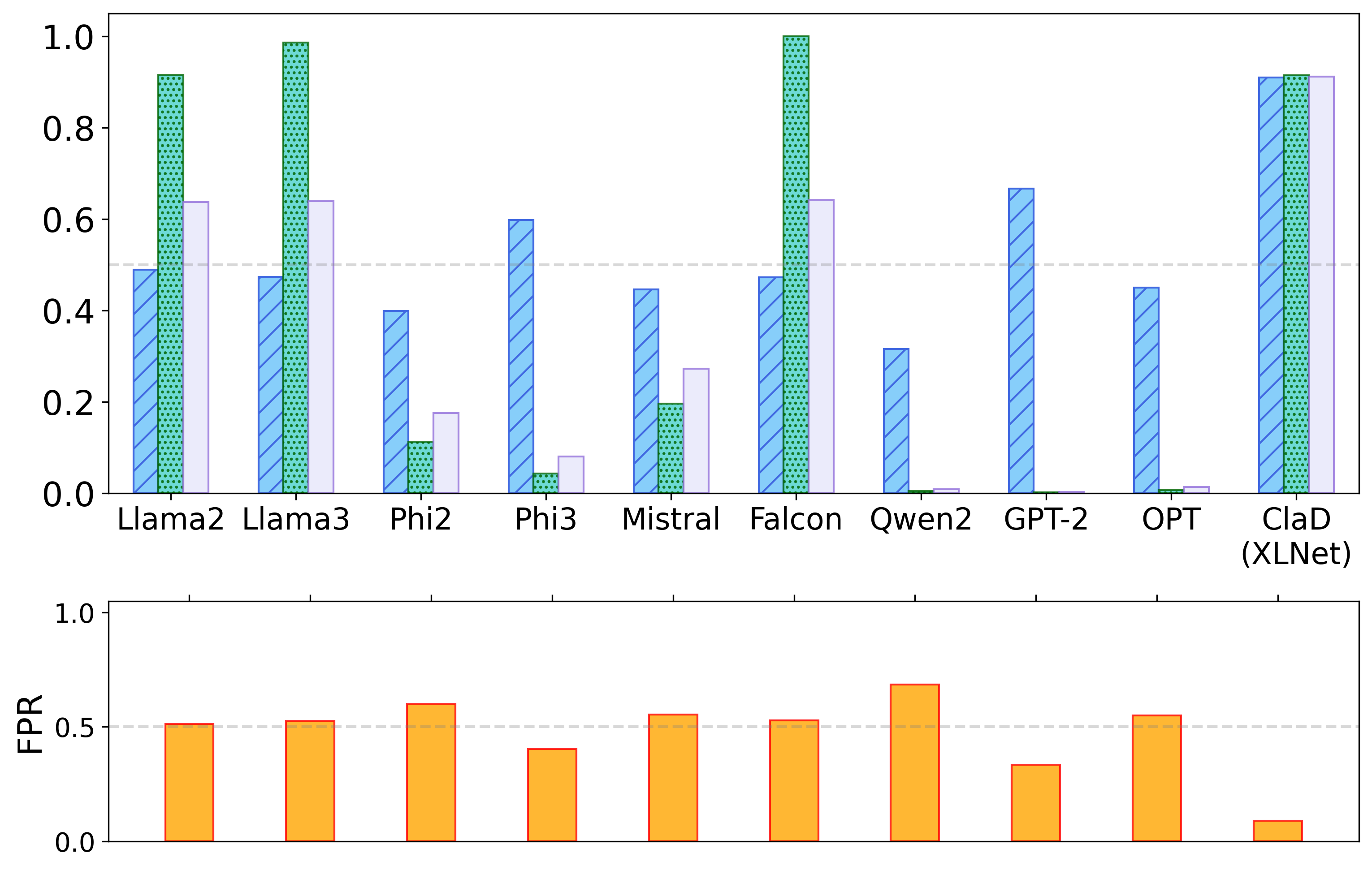}
  \caption{Figurative language: sarcasm}
  \end{subfigure}%
    ~ 
\begin{subfigure}[t]{0.33\textwidth}
  \centering
  \includegraphics[width=0.98\textwidth]{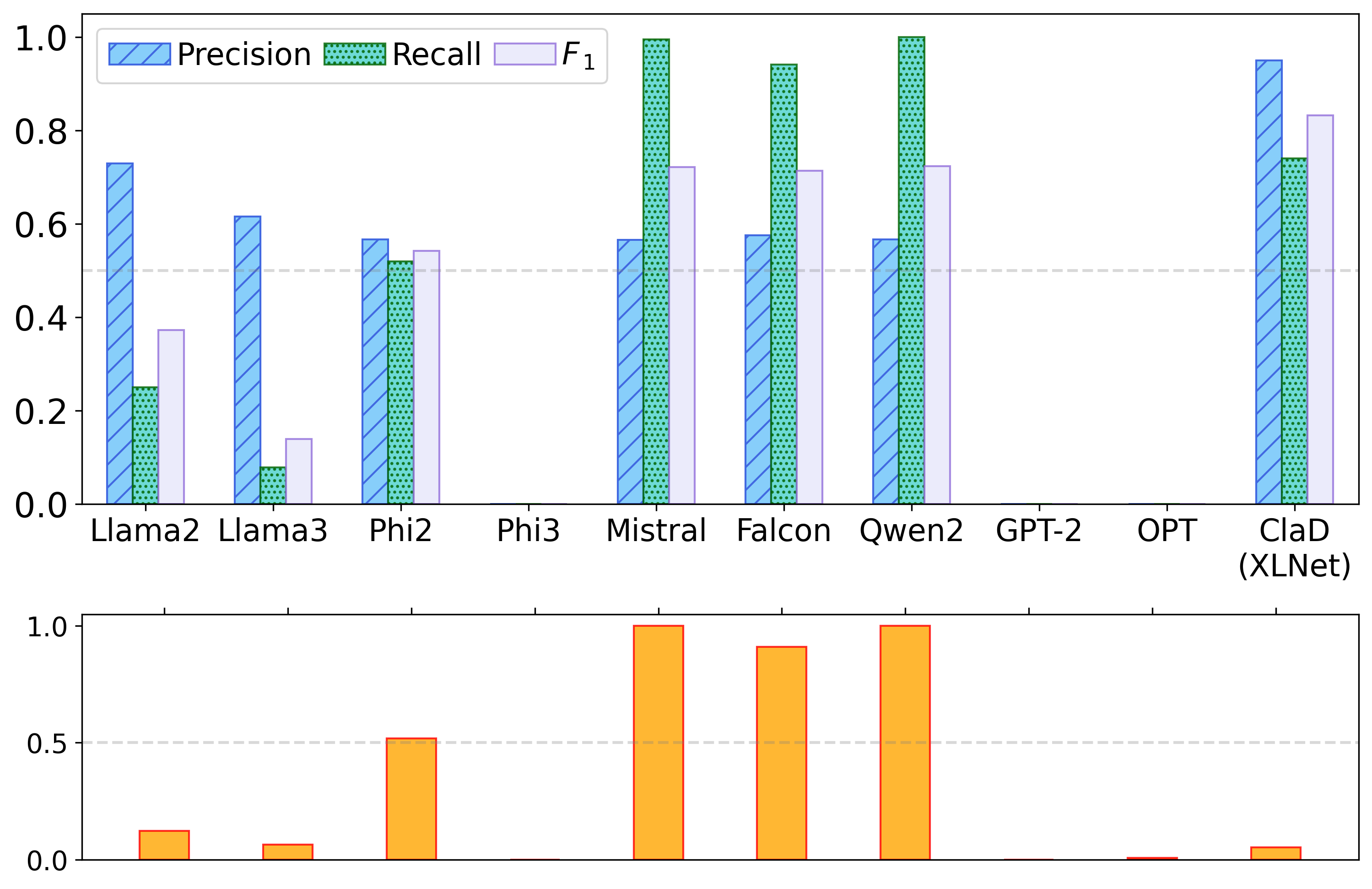}
  \caption{Figurative language: metaphors}
\end{subfigure}%
    ~ 
\begin{subfigure}[t]{0.33\textwidth}
  \centering
  \includegraphics[width=0.98\textwidth]{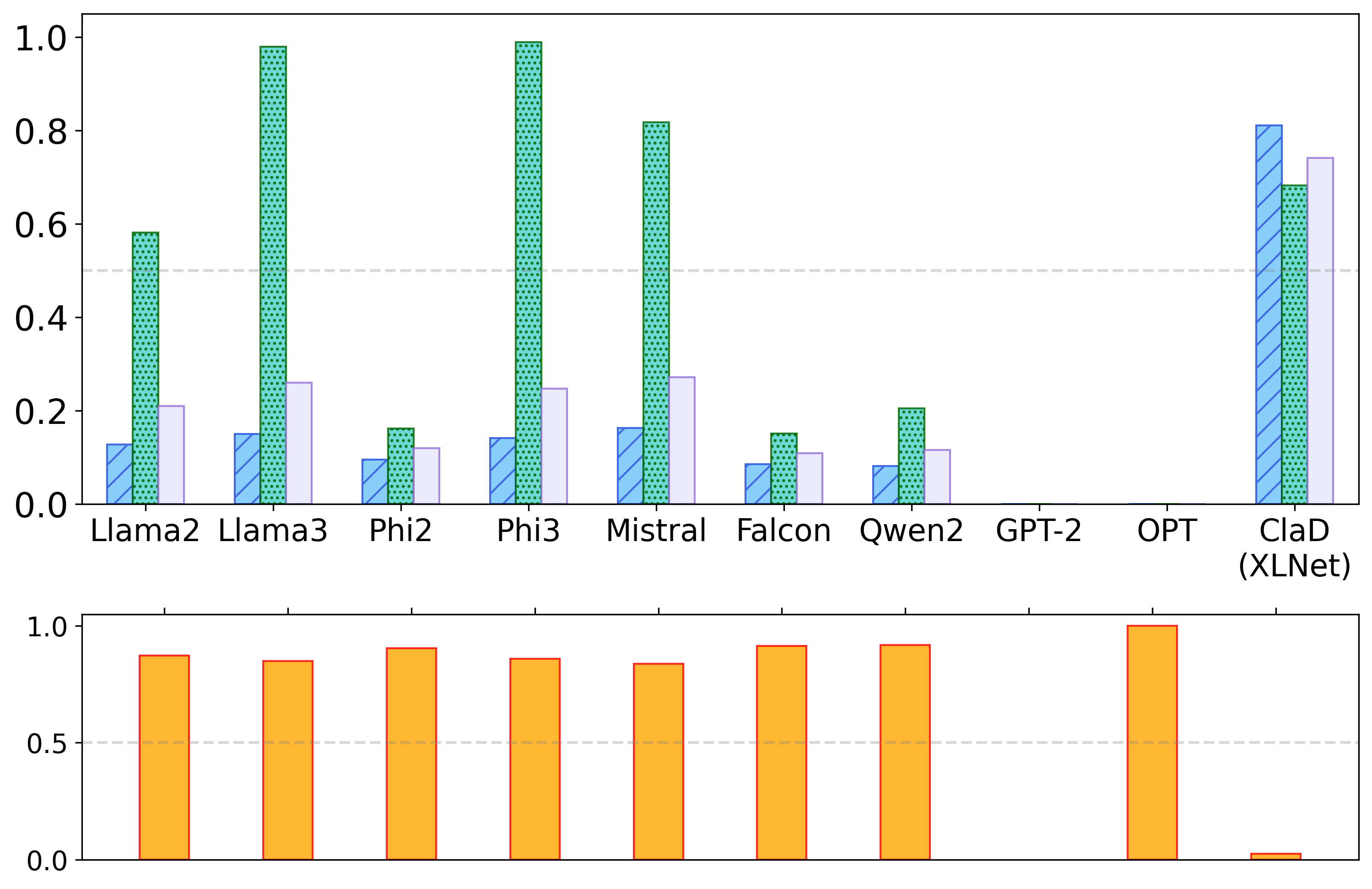}
  \caption{Deviant language: sexism}
\end{subfigure}
\caption{Comparison of zero-shot evaluation of a suite of nine large language models (left to right): Llama2, Llama3, Phi2, Phi3, Mistral-7B, Falcon, Qwen2, GPT-2, and OPT, against ClaD's single-epoch training (rightmost).}
\label{fig:zero-shot-llm-results}
\end{figure*}

\paragraph{Few-shot and Low-resource Experiments:}
We conducted fine-tuning experiments for the classification models under both few-shot and low-resource scenarios. In the few-shot setting, we used 5 training samples, while in the low-resource setting, we used 100 training samples. Training was conducted up to 10 epochs. We selected a variety of mainstream LLMs: Falcon, GPT-2, Llama2, Llama3, Mistral-7B, OPT, Phi2, Phi3, and Qwen2. These models are loaded via the \texttt{AutoModelForSequenceClassification} module provided by HuggingFace Transformers. To train the models with limited computational resources, we employed 4-bit quantization (e.g., \texttt{nf4}) in conjunction with Low-Rank Adaptation (LoRA)~\cite{hu2022lora} for efficient parameter tuning. Specifically, we set the LoRA rank (\texttt{r}) to 16 and the LoRA scaling factor (\texttt{lora\_alpha}) to 8 (32 for GPT-2), with a dropout rate (\texttt{lora\_dropout}) of 0.05 (0.1 for GPT-2). For optimization, HuggingFace \texttt{Trainer} was used with its default settings, with cross-entropy loss adopted for binary classification. The learning rate was set to $5 \times 10^{-5}$, the weight decay set to 0.01, and the batch sizes were 24 for training and 6 for validation (per device).


\begin{figure*}[!htp]
\centering
\begin{subfigure}[t]{.33\textwidth}
    \centering
    \includegraphics[width=\textwidth]{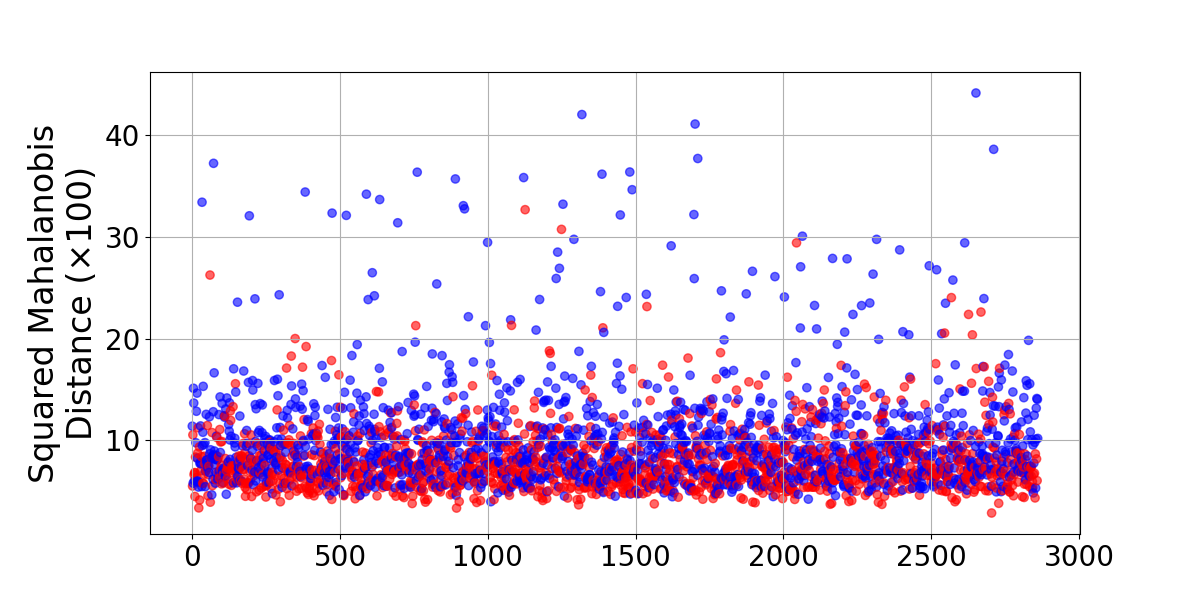}
    \vspace{6pt}
    \includegraphics[width=\textwidth]{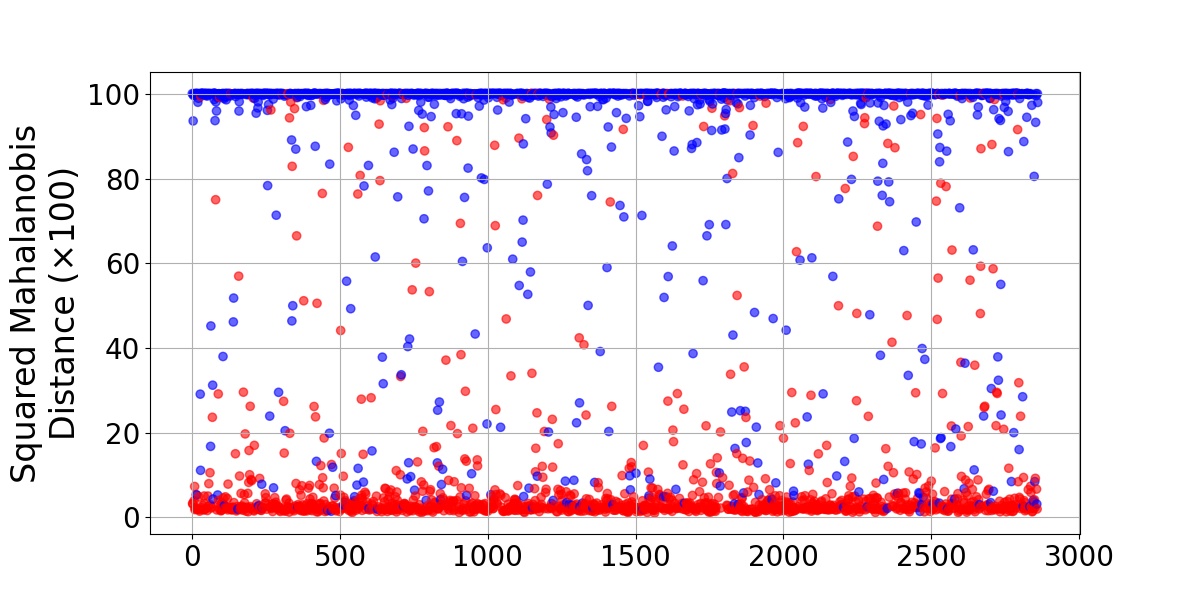}
    \caption{Sarcasm detection (\textit{SH})}
\end{subfigure}%
\hfill
\begin{subfigure}[t]{.33\textwidth}
    \centering
    \includegraphics[width=\textwidth]{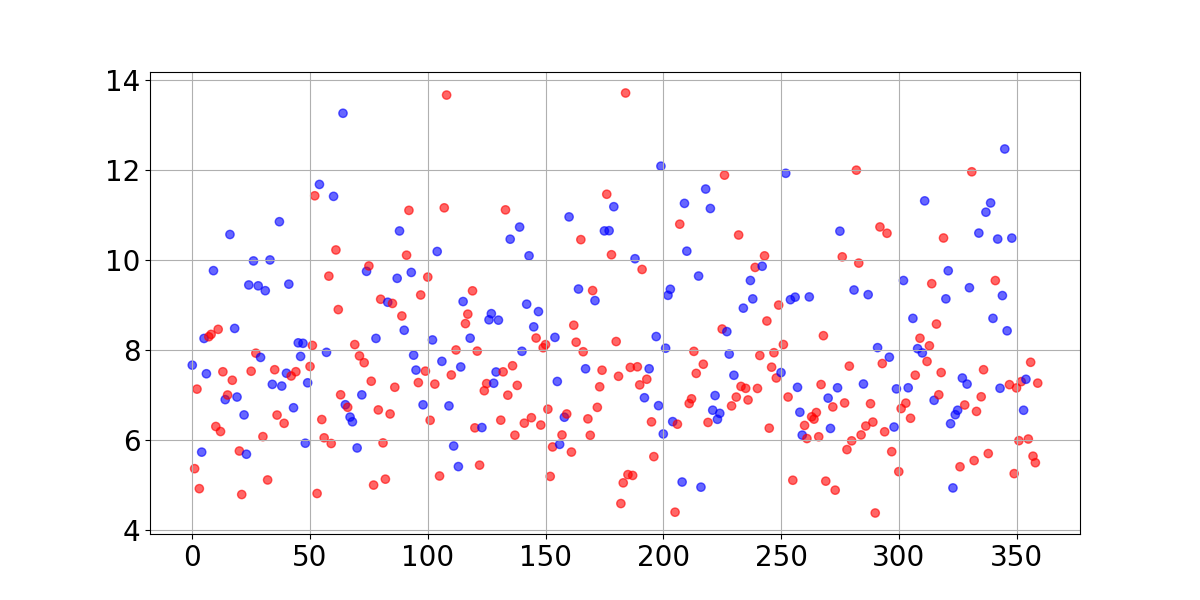}
    \vspace{6pt}
    \includegraphics[width=\textwidth]{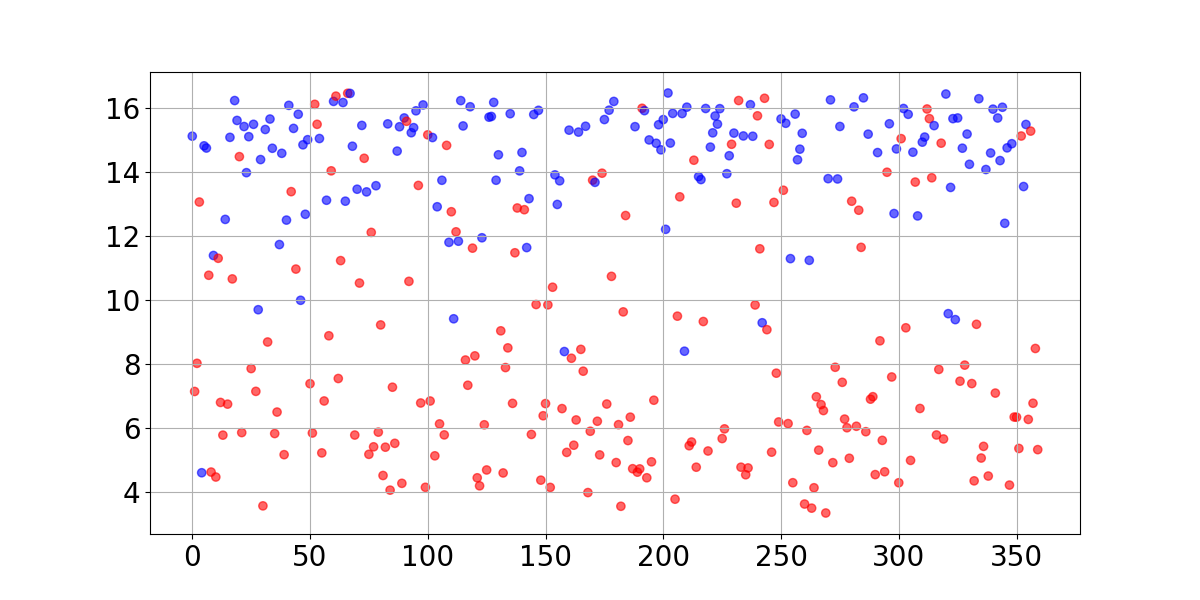}
    \caption{Metaphor detection (\textit{TroFi)})}
\end{subfigure}%
\hfill
\begin{subfigure}[t]{.33\textwidth}
    \centering
    \includegraphics[width=\textwidth]{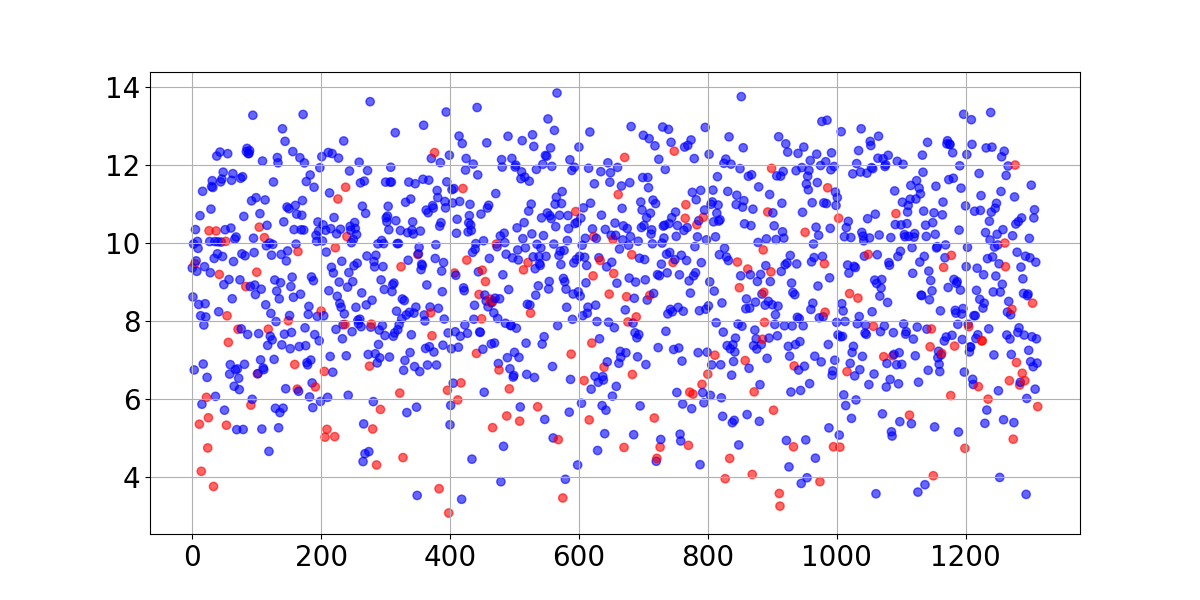}
    \vspace{6pt}
    \includegraphics[width=\textwidth]{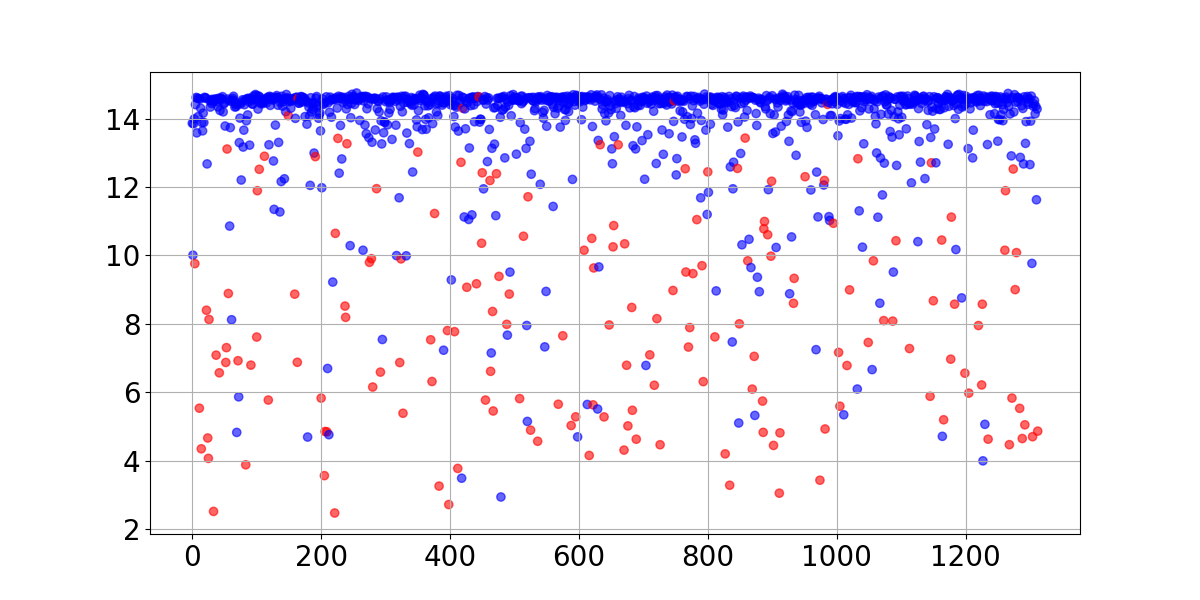}
    \caption{Sexism detection (\textit{CMSB})}
\end{subfigure}%
\caption{Scatter plots of squared Mahalanobis distance for three test datasets before and after training. Red and blue indicate the target and non-target classes, respectively.}
\vspace{-11pt}
\label{fig:separability}
\end{figure*}

\paragraph{Zero-shot Experiments:} The Transformer library's \verb|AutoModelForSequenceClassification| API is used. This API automatically adds a linear classification head on top of the model's pooled output, enabling it to be handled as logits for classification tasks. Specifically, for binary classification, the output consists of 2D logits, representing the likelihood of the input belonging to either class. This approach focuses on \textit{classification}, as opposed to letting a LLM \textit{generate} its answer (e.g., \verb|1| or \verb|0|) in response. To guide the classification, the following prompt is prepended to every input:

{
\renewenvironment{quote}
  {\list{}{\leftmargin=2mm \rightmargin=2mm} \item }
  {\endlist}
{\small
\begin{quote}
\texttt{Please identify if the following text is an example of <task-word>. Reply with 1 if it exhibits <task-word>, and 0 otherwise:} \\
\texttt{<input sentence>}
\end{quote}
}}
where the placeholder \verb|<task-word>| is replaced by the specific task of interest (i.e., sarcasm, metaphor, sexism), and \texttt{<input sentence>} is the text being classified. This enables the model to classify the input sentence based on the particular task while maintaining the flexibility to adapt to various tasks by simply changing the task-word.

\paragraph{Class Distillation Experiments:}
We used a sliding window mechanism (window size: 100 × batch size for small datasets, and expanded to 500 × batch size for large datasets, update frequency: batch size) to efficiently update the Mahalanobis distance parameters (mean and covariance matrix) during training by incrementally processing batch data and computing statistics using the latest model-generated embeddings, dynamically adapting the parameters while maintaining computational efficiency. We use batch sizes of 16 for the sarcasm detection and metaphor tasks, and 40 for the sexism detection task.

\section{Zero-shot classification}
\label{app:zero-shot}

Converting LLM logits to probabilities via softmax, we observe that
ClaD's single-epoch training substantially outperforms zero-shot predictions (shown in \autoref{fig:zero-shot-llm-results}. While this is perhaps unsurprising, the magnitude of improvement is notable: $F_1$ score improvements range from +27.0\% (vs. Falcon) to +90.9\% (vs. GPT-2) on \textit{SH}, +10.8\% (vs. Qwen2) to +83.2\% (vs. OPT) on \textit{TroFi}, and +46.9\% (vs. Mistral) to +74.1\% (vs. OPT) on \textit{CMSB}. In sarcasm detection, LLMs perform near-randomly (accuracies: 0.47-0.53, AUC $\approx$ 0.5), with models either aggressively over-predicting (Falcon, Llama2, Llama3: recall $\geq$ 0.91, but very low precision) or being overly conservative (GPT-2, Phi3: extremely low recall with moderate precision). Metaphor detection shows modest improvements (accuracies: 0.43-0.57, max. AUC: 0.56), though extreme behaviors persist: Falcon, Qwen2 and Mistral favor recall over precision, while Llama2 do the opposite. Sexism detection reveals poor adaptation to class imbalance, with extremely high FPR ($\geq$ 83\%) across all LLMs. Some models also exhibit task-specific inconsistencies, such as GPT-2 alternating between conservative and aggressive predictions in sarcasm and metaphor detection, respectively.

\section{Comparing Anomaly Detection Methods}
\label{app:ablation}
\begingroup
\renewcommand{\arraystretch}{1.25}
\setlength{\tabcolsep}{3pt}
\begin{table}[!t]
\small
\centering
\begin{tabularx}{\linewidth}{@{}l
    >{\raggedleft\arraybackslash}X
    >{\raggedleft\arraybackslash}X
    >{\raggedleft\arraybackslash}X
    >{\raggedleft\arraybackslash}X}
\toprule
 & Acc & Pr &  FPR & $F_1$ \\
\midrule
\multicolumn{5}{c}{Sarcasm detection on \textit{Sarcasm Headlines}} \\
Mahalanobis $\beta$-decision & 0.580 & 0.557 & 0.100 & 0.547 \\
One-class SVM                & 0.494 & 0.465 & 0.494 & 0.473 \\
Isolation Forest             & 0.472 & 0.472 & 0.998 & 0.641 \\
Autoencoder                  & 0.530 & 0.517 & 0.046& 0.099\medskip\\
\multicolumn{5}{c}{Metaphor detection on \textit{TroFi}} \\
Mahalanobis $\beta$-decision & 0.678 & 0.680 & 0.500 & 0.741 \\
One-class SVM                & 0.614 & 0.602 & 0.814 & 0.734 \\
Isolation Forest             & 0.614 & 0.610 & 0.737 & 0.721 \\
Autoencoder                  & 0.578 & 0.575 & 0.942 & 0.724\medskip\\
\multicolumn{5}{c}{Sexism detection on \textit{Call Me Sexist But}} \\
Mahalanobis $\beta$-decision & 0.134 & 0.134 & 1.000 & 0.236 \\
One-class SVM                & 0.602 & 0.141 & 0.364 & 0.206 \\
Isolation Forest             & 0.135 & 0.133 & 0.996 & 0.234 \\
Autoencoder                  & 0.827 & 0.121 & 0.051 & 0.066\smallskip\\
\bottomrule
\end{tabularx}
\caption{Comparing Mahalanobis $\beta$-decision (Algorithm~\ref{alg:decision}) and standard outlier detection methods (One-class SVM, Isolation Forest, and Autoencoder), demonstrating that the former achieves higher $F_1$ scores and lower false positive rates (FPR) across most tasks.}
\vspace{-11pt}
\label{tab:ablation-2}
\end{table}
\endgroup
Additionally, \autoref{tab:ablation-2} shows that treating deviant or figurative language merely as ``out of the ordinary'' is insufficient. We compare our $\beta$-decision algorithm against traditional anomaly detection methods like isolation forest (IF)~\cite{liu2008isolation, liu2010detecting}, autoencoders~\cite{chalapathy2019deep}, and one-class SVM~\cite{noumir2012occ} -- all on the same pretrained model -- and further demonstrate the necessity of an inference algorithm based on an understanding of the target class manifold. Our novel decision algorithm (Algorithm \ref{alg:decision}) achieves superior $F_1$ scores across all tasks, with one exception: IS performs better on sarcasm detection. However, this happens at the expense of significantly higher FPR and lower target-class precision.



\section{Mahalanobis Contrast and Separability}
\label{app:scatter-plots}
Training with our Mahalanobis mean loss function, $\mathcal{L}_{\textsc{mah},\mu}$ (\autoref{eq:mah_mean_loss}) has a significant impact on classification, evident in the results presented in our ablation experiments~\autoref{ssec:ablation}. Here, we add visualizations of class separability, using the squared Mahalanobis distance (\autoref{eq:sq-mah_distance}). \autoref{fig:separability} presents scatter plots of squared Mahalanobis distance for the three test datasets before and after training, where the target and non-target class instances are shown in red and blue, respectively. It is clearly demonstrated that training with our Mahalanobis loss function leads to a distinct increase in class separability.



\end{document}